\documentclass[journal]{IEEEtran}
\usepackage{blindtext}
\usepackage{cite}
\usepackage[pdftex]{graphicx}
\DeclareGraphicsExtensions{.pdf,.jpeg,.png}
\usepackage{algorithm}
\usepackage[noend]{algpseudocode}
\usepackage[tight,footnotesize]{subfigure}
\usepackage{caption}
\usepackage{color}
\usepackage[font=footnotesize]{subfig}
\usepackage{url}
\usepackage{amssymb}
\usepackage{amsmath}
\usepackage{makecell}
\usepackage{xcolor}

\newtheorem{theorem}{Theorem}
\newtheorem{corollary}{Corollary}[theorem]

\newcommand{\R}{\mathbb{R}}
\newcommand{\N}{\mathbb{N}}

\newcommand{\unit}{1\!\!1}

\makeatletter
\def\BState{\State\hskip-\ALG@thistlm}
\makeatother

\begin{document}
\title{Compact and Computationally Efficient Representation of Deep Neural Networks}
\author{Simon Wiedemann, Klaus-Robert M{\"u}ller$^*$,~\IEEEmembership{Member,~IEEE}, and Wojciech Samek$^*$,~\IEEEmembership{Member,~IEEE}
\thanks{This work was supported by the Fraunhofer Society through the MPI-FhG collaboration project ``Theory \& Practice for Reduced Learning Machines''. This research was also supported by the German Ministry for Education through the Berlin Big Data Center under Grant 01IS14013A and the Berlin Center for Machine Learning under Grant 01IS18037I, and by the Institute for Information \& Communications Technology Promotion and funded by the Korea government (MSIT) (No. 2017-0-01779 and No. 2017-0-00451). }
\thanks{S. Wiedemann and W. Samek are with Fraunhofer Heinrich Hertz Institute, 10587 Berlin, Germany (e-mail: wojciech.samek@hhi.fraunhofer.de)}
\thanks{K.-R. M{\"u}ller is with the Technische Universit{\"a}t Berlin, 10587 Berlin, Germany, with the Max Planck Institute for Informatics, 66123 Saarbr{\"u}cken, Germany, and also with the Department of Brain and Cognitive Engineering, Korea University, Seoul 136-713, South Korea (e-mail: klaus-robert.mueller@tu-berlin.de)}}
\markboth{Wiedemann et al. - Efficient Representations of Deep Neural Networks} 
{Wiedemann et al. - Efficient Representations of Deep Neural Networks} 

\maketitle

\begin{abstract}
At the core of any inference procedure in deep neural networks are dot product operations, which are the component that require the highest computational resources.
For instance, deep neural networks such as VGG-16 require up to 15 giga-operations in order to perform the dot products present in a single forward pass, which results in significant energy consumption and therefore limit their use in resource-limited environments, e.g., on embedded devices or smartphones. A common approach to reduce the cost of inference is to reduce its memory complexity by lowering the entropy of the weight matrices of the neural network, e.g., by pruning and quantizing their elements. However, the quantized weight matrices are then usually represented either by a dense or sparse matrix storage format, whose associated dot product complexity is \textit{not} bounded by the entropy of the matrix. This means that the associated inference complexity ultimately depends on the implicit statistical assumptions that these matrix representations make about the weight distribution, which can be in many cases suboptimal. 
In this paper we address this issue and present new efficient representations for matrices with low entropy statistics. These new matrix formats have the novel property that their memory \textit{and} algorithmic complexity are implicitly bounded by the entropy of the matrix, consequently implying that they are guaranteed to become more efficient as the entropy of the matrix is being reduced. In our experiments we show that performing the dot product under these new matrix formats can indeed be more energy and time efficient under practically relevant assumptions. For instance, we are able to attain up to x42 compression ratios, x5 speed ups and x90 energy savings when we convert in a \textit{lossless} manner the weight matrices of state-of-the-art networks such as AlexNet, VGG-16, ResNet152 and DenseNet into the new matrix formats and benchmark their respective dot product operation.
\end{abstract}


\IEEEpeerreviewmaketitle

\section{Introduction}
The dot product operation between matrices constitutes one of the core operations in almost any field in science. 
Examples are the computation of approximate solutions of complex system behaviors in physics \cite{CompPhys_survey}, iterative solvers in mathematics \cite{Num_Math_survey} and features in computer vision applications \cite{CV_feature_extraction_survey}.
Also deep neural networks heavily rely on dot product operations in their inference \cite{DL_book, DL_tricks_trade}; e.g., networks such as VGG-16 require up to 16 dot product operations, which results in 15 giga-operations for a single forward pass.
Hence, lowering the algorithmic complexity of these operations and thus increasing their efficiency is of major interest for many modern applications. Since the complexity depends on the data structure used for representing the elements of the matrices, a great amount of research has focused on designing data structures and respective algorithms that can perform efficient dot product operations \cite{Matrix_mul_survey, Fast_matrix_mul, sparse_matrices_research}.

Of particular interest are the so called \textit{sparse matrices}, a special type of matrices that have the property that many of their elements are zero valued. In principle, one can design efficient representations of sparse matrices  by leveraging the prior assumption that most of their element values are zero and therefore, only store the non-zero entries of the matrix. Consequently, their storage requirements become of the order of the number of non-zero values. However, having an efficient representation with regard to storage requirement does not imply that the dot product algorithm associated to that data structure will also be efficient. Hence, a great part of the research was focused on the design of data structures that  have as well low complex dot product algorithms \cite{sparse_matrices_research, CSX, DCSR, fast_sparse_matrix_mul}. 
However, by assuming sparsity alone we are implicitly imposing a spike-and-slab prior\footnote{That is, a delta function at 0 and a uniform distribution over the non-zero elements.} over the probability mass distribution of the elements of the matrix. If the actual distribution of the elements greatly differs from this assumption, then the data structures devised for sparse matrices become inefficient. Hence, sparsity can be a too constrained assumption for some applications of current interest, e.g., representation of quantized neural networks.  

In this work, we alleviate the shortcomings of sparse representations by considering a more relaxed prior over the distribution of the matrix elements. More precisely, we assume that the empirical probability mass distribution of the matrix elements has a low entropy value as defined by Shannon \cite{Shannon}. 
Mathematically, sparsity can be considered a subclass of the general family of low entropic distributions. In fact, sparsity measures the min-entropy of the element distribution, which is related to Shannon's entropy measure through Renyi's generalized entropy definition \cite{renyi1961}.
With this goal in mind, we ask the question: 
\begin{quote}
\it ``Can we devise efficient data structures under the implicit assumption that the entropy of the distribution of the matrix elements is low?''
\end{quote}

We want to stress once more that by efficiency we regard two related but distinct aspects
\begin{enumerate}
\item efficiency with regard to storage requirements
\item efficiency with regard to algorithmic complexity of the dot product associated to the representation
\end{enumerate}
For the later, we focus on the number of elementary operations required in the algorithm, since they are related to the energy and time complexity of the algorithm.
It is well known that the minimal bit-length of a data representation is  bounded by the entropy of it's distribution \cite{Shannon}. Hence, matrices with low entropic distributions automatically imply that we can design data structures that do not require high storage resources. 
In addition, as we will discuss in the next sections, low entropic distributions also attain gains in efficiency if these data structures implicitly encode the distributive law of multiplications. By doing so, a great part of the algorithmic complexity of the dot product is reduced to the order of the number of  shared weights per row in a matrix. This number is related to the entropy, such that it is small as long as the entropy of the matrix is low. Therefore, these data structures not only attain higher compression gains, but also require less total number of operations when performing the dot product. 

Our contributions can be summarized as follows:
\begin{itemize}
\item We propose new highly efficient data structures that exploit on the prior that the matrix has a low number of shared weights per row (i.e., low entropy).
\item We provide a detailed analysis of the storage requirements and algorithmic complexity of performing the dot product associated to these data structures.
\item We establish a relation between the known sparse and the proposed data structures. Namely, sparse matrices belong to the same family of low entropic distributions, however, they can be considered a more constrained subclass of them.
\item We show through experiments that indeed, these data structures attain gains in efficiency on simulated as well as real-world data. In particular, we show that up to x42 compression ratios, x5 speed ups and x90 energy savings can be achieved when we benchmark the compressed weight matrices of state-of-the-art neural networks relative to the matrix-vector multiplication.
\end{itemize}

In the following Section \ref{sec:efficientNN} we introduce the problem of efficient representation of neural networks and briefly review related literature.
In Section \ref{sec: A simple but conceptual example} the proposed data structures are given. We demonstrate through a simple example that these data structures are able to: 1) achieve higher compression ratios than their respective dense and sparse counterparts and 2) reduce the algorithmic complexity of performing the dot product. Section \ref{sec: A list of crude data structures} analyses the storage and energy complexity of these novel data structures. Experimental evaluation is performed in Section \ref{sec: Experiments} using simulations as well as state-of-the-art neural networks such as AlexNet, VGG-16, ResNet152 and DenseNet. Section \ref{sec:conclusion} concludes the paper with a discussion. 

\section{Efficient Inference in Neural Networks}
\label{sec:efficientNN}
Deep neural networks \cite{DeepLearning, Schmidhuber14_DL_survey} became the state-of-the-art in many fields of machine learning, such as in computer vision, speech recognition, natural language processing \cite{he2015, BosTIP18, 2016arXiv161000087D, bahdanau2014}, and have been progressively also used in the sciences, e.g. physics \cite{DL_physics}, neuroscience \cite{DL_EEG_Wojciech}, chemistry \cite{Schuett2017, Chmielae1603015}. In their most basic form, they constitute a chain of affine transformations concatenated with a non linear function which is applied element-wise to the output. Hence, the goal is to learn the values of those transformation or weight matrices (i.e., parameters) such that the neural network performs it's task particularly well. The procedure of calculating the output prediction of the network for a particular input is called {\it inference}. 
The computational cost of performing inference is dominated by computing the affine transformations (thus, the dot products between matrices). Since today's neural networks perform many dot product operations between large matrices, this greatly complicates their deployment onto resource constrained devices.

However, it has been extensively shown that most neural networks are overparameterized, meaning that there are many more parameters than actually needed for the tasks of interest \cite{DLC_low_rank, NN_distillation, deep_compression, VD_sparsifies}. This implies that these networks are highly inefficient with regard to the resources they require when performing inference. This fact motivated an entire research field of model compression \cite{DLC_survey}. One of the suggested approaches is to: 1) compress the weight elements of the neural network without (considerably) affecting their prediction accuracy and 2) convert the resulting weights into a representation that achieves high compression ratios and is able to execute the dot product operation efficiently. Whilst there has been a plethora of work focusing on the first step \cite{Opt_brain_damage, Opt_brain_surgeon, Learning_Weight_and_Connections, VD_sparsifies, Google_FP, FP_CNN_Qualcomm, Terniray_weights, towards_limit_quantization_DNN, Universal_dnn_compression, deep_compression, soft_weight_sharing, BayesianCompression, improved_BayesianCompression}, previous literature has not focused as much on the second part. As a consequence, most of the research has focused on developing techniques that either sparsify the networks weights \cite{Opt_brain_damage, Opt_brain_surgeon, Learning_Weight_and_Connections, VD_sparsifies} or reduce the cardinality of the weight elements \cite{Google_FP, FP_CNN_Qualcomm, Terniray_weights}, since then sparse matrix representations or dense matrices with compressed numerical representations can be employed in order to efficiently perform inference.

However, this greatly reduces the possible efficiency gains that can be achieved. In fact, highest reported compression gains are attained with techniques that either implicitly \cite{deep_compression, BayesianCompression} or explicitly \cite{towards_limit_quantization_DNN, Universal_dnn_compression, soft_weight_sharing, improved_BayesianCompression} attempt to reduce the entropy of the weight matrices of the network. To recall, throughout this work we consider the entropy of the empirical probability mass distribution of the weight elements. That is, we first identify the set of unique elements that appear in the matrix, denoted as $\Omega$. Then, for each element in $\omega_k \in \Omega$, we count it's frequency of appearance and divide it by the total number of elements in the matrix, resulting in the probability mass value $p_k = \#(\omega_k)/N$, where $\#(\cdot)$ is the counting operator and $N$ the total number of elements in the matrix. Finally, we calculate Shannon's entropy $H = -\sum_k p_k\log_2p_k$.

However, with no other means for representing the resulting compressed weight matrices, the achievable efficiency gains are bounded by the limitations of the sparse or dense representations.

\begin{figure}[t]
\includegraphics[width=\columnwidth,clip,keepaspectratio]{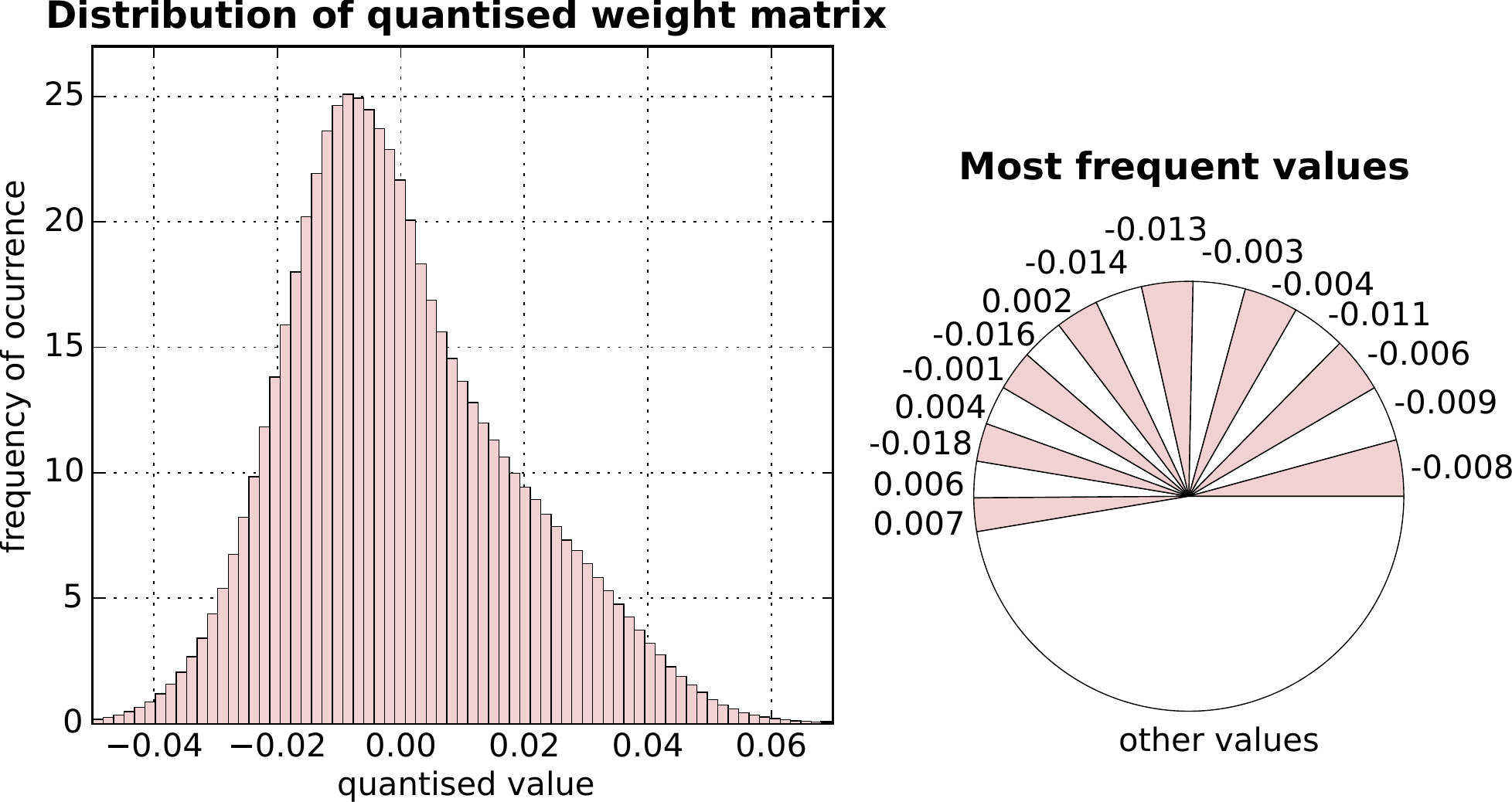}
\caption{Distribution of the weight matrix of the last layer of the VGG-16 neural network \cite{VGG} after quantization. The respective matrix is $1000\times 4096$ dimensional, transforming the 4096 last-layer features onto 1000 output classes. We applied an uniform quantizer over the range of values, with $2^7$ quantization points, which resulted in no loss of accuracy on the classification task. \underline{Left}: Probability mass distribution. \underline{Right}: Frequency of appearance of the 15 most frequent values.}
\label{Fig: VGG class distr}
\end{figure}

For instance, figure \ref{Fig: VGG class distr} demonstrates the discrepancy between the sparsity assumption and the real distribution of weight elements.
It plots the distribution of the weight elements of the last classification layer of VGG-16 \cite{VGG} ($1000\times 4096$ dimensional matrix), after having applied uniform quantization on the weight elements. We stress that the prediction accuracy and generalization of the network was not affected by this operation. On the one hand, as we can see, the distribution of the compressed layer does not satisfy the sparsity assumption, i.e., there is not one particular element (such as 0) that appears specially frequent in the matrix. The most frequent value is -0.008 and it's frequency of appearance does not dominate over the others (about 4.2\%).
On the other hand, naively compressing the numerical values of the matrix elements down to a trivial 7-bit representation would also result in an inefficient representation. Since the activation values are still represented in single precision floating point values\footnote{In this case, compressing the activation values down to a 7-bit representation would have significantly harmed the prediction accuracy of the network.}, the respective dot product algorithm would require multiple, mostly expensive decoding operations in order to convert back each element of the weight matrix into it's original 32-bit floating point value.

Hence, neither sparse matrix representations nor the (compressed) dense representations can efficiently exploit the statistical properties of the weight matrix.

In this work, we overcome these limitations and present new matrix representations that become more efficient as the entropy of the weight matrices is reduced. In particular, their complexity depend partially on the number of shared weights present in the matrix, which is reduced as the entropy of the matrix is reduced. Indeed, we notice that for the matrix in figure \ref{Fig: VGG class distr} most of the entries are dominated by only 15 distinct values, which is 1.5\% of the number of columns of the matrix. In the next section we will describe with a simple example how these new representations leverage on this property in order to achieve both, high compression ratios and efficient dot products. 

\section{Data structures for matrices with low entropy statistics}
\label{sec: A simple but conceptual example}
In this section we introduce the proposed data structures and show that they implicitly encode the distributive law.
Consider the following matrix

\begin{align}
\setcounter{MaxMatrixCols}{20}
  M = \begin{pmatrix}
 0 & 3 & 0 & 2 & 4 & 0 & 0 & 2 & 3 & 4 & 0 & 4 \\
 4 & 4 & 0 & 0 & 0 & 4 & 0 & 0 & 4 & 4 & 0 & 4 \\
 4 & 0 & 3 & 4 & 0 & 0 & 0 & 4 & 0 & 2 & 0 & 0 \\
 0 & 0 & 0 & 4 & 4 & 4 & 0 & 3 & 4 & 4 & 0 & 0 \\
 0 & 4 & 4 & 0 & 0 & 4 & 0 & 4 & 0 & 0 & 0 & 0 
  \end{pmatrix}
  \nonumber
\end{align}
Now assume that we want to: 1) store this matrix with the minimum amount of bits and 2) perform the dot product with a vector $a\in \R^{12}$ with the minimum complexity. 

\subsection{Minimum storage}
We firstly comment on the storage requirement of dense and sparse formats and then introduce two new formats which more effectively store matrix $M$.\\[+6px]
\noindent{\underline{Dense format}:
Arguably the simplest way to store the matrix $M$ is in it's so called dense representation. That is, we store it's elements in a $5\times 12$ long array  (in addition to it's dimensions $m=5$ and $n=12$).\\[+6px]
\noindent{\underline{Sparse format}}:
However, notice that more than 50\% of the entries are 0. Hence, we may be able to attain a more compressed representation of this matrix if we store it in one of the well known sparse data structure, for instance, in the \textit{Compressed Sparse Row} (or CSR in short) format. This particular format stores the values of the matrix in the following way:
\begin{itemize}
\item Scans the non-zero elements in row-major order (that is, from left to right, up to down) and stores them in an array (which we denote as $W$).
\item Simultaneously, it stores the respective column indices in another array (which we call $colI$).
\item Finally, it stores pointers that signal when a new row starts (we denote this array as $rowPtr$).
\end{itemize}
Hence, our matrix $M$ would take the form
\begin{align}
W: & [3, 2, 4, 2, 3, 4, 4, 4, 4, 4, 4, 4, 4, 4, 3, \nonumber \\
   &  4, 4, 2, 4, 4, 4, 3, 4, 4, 4, 4, 4, 4] \nonumber \\
colI: & [ 1,  3,  4,  7,  8,  9, 11,  0,  1,  5,  8,  9, 11,  0, \nonumber \\
      & 2,  3,  7, 9,  3,  4,  5,  7,  8,  9,  1,  2,  5,  7] \nonumber \\
rowPtr: & [ 0,  7, 13, 18, 24, 28]
\nonumber
\end{align} 
If we assume the same bit-size per element for all arrays, then the CSR data structure does not attain higher compression gains in spite of not saving the zero valued elements (62 entries vs.\ 60 that are being required by the dense data structure).\\[+6px]
We can improve this by exploiting the low-entropy property of matrix $M$.
In the following, we propose two new formats which realize this.\\[+6px]
\noindent\underline{Compressed Entropy Row (CER) format}:
Firstly, notice that many elements in $M$ share the same value. In fact, only the four values $\Omega=\{0,4,3,2\}$ appear in the entire matrix. Hence, it appears reasonable to assume that data structures that repeatedly store these values (such as the dense or CSR structures) induce high redundancies in their representation. Therefore, we propose a data structure where we only store those values once. Secondly, notice that different elements appear more frequent than others, and their relative order does not change throughout the rows of the matrix. Concretely, we have a set of unique elements $\Omega=\{0,4,3,2\}$ which appear $P_{\#} = \{32, 21, 4, 3\}$ times respectively in the matrix, and we obtain the same relative order of highest to lowest frequent value throughout the rows of the matrix. Hence, we can design an efficient data structure which leverages on both properties in the following way:
\begin{enumerate}
\item Store unique elements present in the matrix in an array in frequency-major order (that is, from most to least frequent). We name this array $\Omega$.
\item Store respectively the column indices in row-major order, excluding the first element (thus excluding the most frequent element). We denote it as $colI$.
\item Store pointers that signal when the positions of the next new element in $\Omega$ starts. We name it $\Omega Ptr$. If a particular pointer in $\Omega Ptr$ is the same as the previous one, this means that the current element is not present in the matrix and we jump to the next element. 
\item Store pointers that signal when a new row starts. We name it $rowPtr$. Here, $rowPtr$ points to entries in $\Omega Ptr$.
\end{enumerate}
Hence, this new data structure represents matrix $M$ as
\begin{align}
\Omega: & [0,4,3,2] \nonumber \\
colI: & [ 4,  9, 11,  1,  8,  3,  7,  0,  1,  5,  8,  9, 11,  0, \nonumber \\
      & 3,  7,  2, 9,  3,  4,  5,  8,  9,  7,  1,  2,  5,  7] \nonumber \\
\Omega Ptr: & [ 0,  3,  5,  7, 13, 16, 17, 18, 23, 24, 28] \nonumber \\
rowPtr: & [ 0,  3,  4,  7,  9, 10]
\nonumber
\end{align}
Notice that we can uniquely reconstruct $M$ from this data structure. We refer to this data structure as the \textit{Compressed Entropy Row} (or CER in short) data structure. One can verify that indeed, the CER data structure only requires 49 entries (instead of 60 or 62) attaining as such a compressed representation of the matrix $M$. 

To summarize, the CER representation is able to attain higher compression gains because it leverages on the following two properties: 1) many matrix elements share the same value and 2) the empirical probability mass distribution of the shared weight elements does not change significantly across rows. \\[+6px]
\noindent\underline{Compressed Shared Elements Row (CSER) format}:
In some cases, it may well be that the probability distribution across rows in a matrix are not similar to each other. Hence, the second assumption in the CER data structure would not apply and we would only be left with the first one. That is, we only know that not many distinct elements appear per row in the matrix or, in other words, that many elements share the same value.
The \textit{compressed shared elements row} (or CSER in short) data structure is a slight extension to the previous CER representation. Here, we add an element pointer array, which signals which element in $\Omega$ the $colI$ indices refer to. We called it $\Omega I$. Thus, $\Omega I$ points to entries in $\Omega$, $\Omega Ptr$ to entries in $colI$ and $rowPtr$ to entries in $\Omega Ptr$. Hence, the above matrix would then be represented as follows
\begin{align}
\Omega: & [0,2,3,4] \nonumber \\
colI: & [ 4,  9, 11,  1,  8,  3,  7,  0,  1,  5,  8,  9, 11,  0, \nonumber \\
      & 3,  7,  2, 9,  3,  4,  5,  8,  9,  7,  1,  2,  5,  7] \nonumber \\
\Omega I: & [3, 2, 1, 3, 3, 2, 1, 3, 2, 3] \nonumber \\
\Omega Ptr: & [ 0,  3,  5,  7, 13, 16, 17, 18, 23, 24, 28] \nonumber \\
rowPtr: & [ 0,  3,  4,  7,  9, 10]\nonumber
\end{align}
Thus, for storing matrix $M$ we require 59 entries, which is still a gain but not a significant one.
Notice, that now the ordering of the elements in $\Omega$ is not important anymore, as long as the $\Omega I$ array is accordingly adjusted. Similarly, the ordering of $\Omega I$ at each row can also be arbitrary, as long as the $\Omega Ptr$ and $colI$ array are accordingly adjusted. 

The relationship between CSER, CER and CSR data structures is described in Section \ref{sec: A list of crude data structures}.

\subsection{Dot product complexity}
We just saw that we can attain gains with regard to compression if we represent the matrix in the CER and CSER data structures. However, we can also devise corresponding dot product algorithms that are more efficient than their dense and sparse counterparts.
As an example, consider only the scalar product between the second row of matrix $M$ with an arbitrary input vector $a = [a_1\; a_2\; \ldots \; a_{12}]^\top$.
In principle, the difference in the algorithmic complexity arises because each data structure implicitly encodes a different expression of the scalar product, namely
\begin{align*}
dense: & \; 4a_1 + 4a_2 + 0a_3 +  0a_4 + 0a_5 + 4a_6\\
       &  + 0a_7 + 0a_8 + 4a_9 + 4a_{10} + 0a_{11} + 4a_{12} \\
CSR: & \; 4a_1 + 4a_2 + 4a_6 + 4a_9 + 4a_{10} + 4a_{12} \\
CER/CSER: & \; 4(a_1 + a_2 + a_6 + a_9 + a_{10} + a_{12})
\end{align*}
For instance, the dot product algorithm associated to the dense format would calculate the above scalar product by
\begin{enumerate}
\item loading $M$ and $a$.
\item calculating $4a_0 + 4a_1 + 0a_2 +  0a_3 + 0a_4 +  4a_5 + 0a_6 + 0a_7 + 4a_8 + 4a_{9} + 0a_{10} + 4a_{11}$.
\end{enumerate}
This requires 24 load (12 for the matrix elements and 12 for the input vector elements), 12 multiply, 11 add and 1 write operations (for writing the result into memory). We purposely omitted the accumulate operation which stores the intermediate values of the multiply-sum operations, since their cost can effectively be associated to the sum operation. Moreover, we only considered read/write operations from and into memory. Hence, this makes 48 operations in total.

In contrast, the dot product algorithm associated with the CSR representation would only multiply-add the non-zero entries. It does so by performing the following steps
\begin{enumerate}
\item Load the subset of $rowPtr$ respective to row 2. Thus, $rowPtr \rightarrow [7, 13]$.
\item Then, load the respective subset of non-zero elements and column indices. Thus, $W \rightarrow [4, 4, 4, 4, 4, 4]$ and $colI \rightarrow  [0,  1,  5,  8,  9, 11]$.
\item Finally, load the subset of elements of $a$ respective to the loaded subset of column indices and subsequently multiply-add them to the loaded subset of $W$. Thus, $a\rightarrow [a_0, a_1, a_5, a_8, a_{10}, a_{11}]$ and calculate $4a_0 + 4a_1 + 4a_5 + 4a_8 + 4a_{9} + 4a_{11}$.
\end{enumerate}
By executing this algorithm we would require 20 load operations (2 from the $rowPtr$ and 6 for the $W$, the $colI$ and the input vector respectively), 6 multiplications, 5 additions and 1 write. In total this dot product algorithm requires 32 operations.

However, we can still see that the above dot product algorithm is inefficient in this case since we constantly multiply by the same element 4. Instead, the dot product algorithm associated to, e.g., the CER data structure, would perform the following steps
\begin{enumerate}
\item Load the subset of $rowPtr$ respective to row 2. Thus, $rowPtr \rightarrow [3, 4]$.
\item Load the corresponding subset in $\Omega Ptr$. Thus, $\Omega Ptr \rightarrow [7, 13]$.
\item For each pair of elements in $\Omega Ptr$, load the respective subset in $colI$ and the element in $\Omega$. Thus, $\Omega \rightarrow [4]$ and $colI \rightarrow [0,  1,  5,  8,  9, 11]$.
\item For each loaded subset of $colI$, perform the sum of the elements of $a$ respective to the loaded $colI$. Thus, $a\rightarrow [a_0, a_1, a_5, a_8, a_{10}, a_{11}]$ and do $a_0 + a_1 + a_5 + a_8 + a_{9} + a_{11} = z$.
\item Subsequently, multiply the sum with the respective element in $\Omega$. Thus, compute $4z$.
\end{enumerate}
A similar algorithm can be devised for the CSER data structure. One can find both pseudocodes in the appendix. The operations required by this algorithm are 17 load operations (2 from $rowPtr$, 2 from $\Omega Ptr$, 1 from $\Omega$, 6 from $colI$ and 6 from $a$), 1 multiplication, 5 additions and 1 write. In total these are 24 operations.\\[+6px]
Hence, we have observed that for the matrix $M$, the CER (and CSER) data structure does not only achieve \textbf{higher compression rates}, but it also attains \textbf{gains in efficiency} with respect to the dot product operation.\\[+6px]
In the next section we give a detailed analysis about the storage requirements needed by the data structures and also the efficiency of the dot product algorithm associated to them. This will help us identify when one type of data structure will attain higher gains than the others.

\section{An analysis of the storage and energy complexity of data structures}
\label{sec: A list of crude data structures}
Without loss of generality, in the following we assume that we aim to encode a particular matrix $M\in \Omega^{n\times m=N}$, where it's elements $M_{ij} = \omega_k \in \Omega$ take values from a finite set of elements $\Omega = \{\omega_0, \omega_1, ..., \omega_{K-1}\}$. Moreover, we assign to each element $\omega_k$ a probability mass value $p_k = \#(\omega_k)/N$, where $\#(\omega_k)$ counts the number of times the element $\omega_k$ appears in the matrix $M$. We denote the respective set of probability mass values $P_{\Omega} = \{p_0, p_1, ..., p_{K-1}\}$. In addition, we assume that each element in $\Omega$ appears at least once in the matrix (thus, $p_k > 0$ for all $k=0, ..., K-1$) and that $\omega_0 = 0$ is the most frequent value in the matrix. Finally, we order the elements in $\Omega$ and $P_{\Omega}$ in probability-major order, that is, $p_0 \geq p_1 \geq ...\geq p_{K-1}$.  

\subsection{Measuring the energy efficiency of the dot product}
This work proposes representations that are efficient with regard to storage requirements as well as their dot product algorithmic complexity. For the latter, we focus on the energy requirements, since we consider it as the most relevant measures for neural network compression. However, exactly measuring the energy of an algorithm is unreliable since it depends on the software implementation and on the hardware the program is running on. Therefore, we will model the energy costs in a way that can easily be adapted across different software implementations as well as hardware architectures.

In the following we model a dot product algorithm by a computational graph, whose nodes can be labeled with one of four elementary operations, namely: 1) a \textit{mul} or multiply operation which takes two numbers as input and outputs their multiplied value, 2) a \textit{sum} or summation operation which takes two values as input and outputs their sum, 3) a \textit{read} operation which reads a particular number from memory and 4) a \textit{write} operation which writes a value into memory. Note, that we do not consider read/write operations from/into low level memory (like caches and registers) that store temporary runtime values, e.g., outputs from summation and/or multiplications, since their cost can be associated to those operations. Now, each of these nodes can be associated with an energy cost. Then, the total energy required for a particular dot product algorithm simply equals the total cost of the nodes in the graph. 

However, the energy cost of each node depends on the hardware architecture and on the bit-size of the values involved in the operation. Hence, in order to make our model flexible with regard to different hardware architectures, we introduce four cost functions $\sigma, \mu, \gamma, \delta: \N \rightarrow \R$, which take as input a bit-size and output the energy cost of performing the operation associated to them\footnote{The \textit{sum} and \textit{mul} operations take two numbers as input and they may have different bit-sizes. Hence in this case, we take the maximum of those as a reference for the bit-sizes involved in the operation.}; $\sigma$ is associated to the \textit{sum} operation,  $\mu$ to the \textit{mul}, $\gamma$ to the \textit{read} and $\delta$ to the \textit{write} operation.

Figure \ref{Fig: dot comp graph} shows the computational graph of a simple dot product algorithm for two 2-dimensional input vectors. This algorithm requires 4 \textit{read} operations, 2 \textit{mul}, 1 \textit{sum} and 1 \textit{write}. Assuming that the bit-size of all numbers is $b \in \N$, we can state that the energy cost of this dot product algorithm would be $E = 1\sigma(b) + 2\mu(b) + 4\gamma(b) + 1\delta(b)$.
Note that similar energy models have been previously proposed \cite{energy_aware_pruning, Eyeriss}. 
In the experimental section we validate the model by comparing it to real energy results measured by previous authors.

Considering this energy model we can now provide a detailed analysis of complexity of the CER and CSER data structure. However, we start with a brief analysis of the storage and energy requirements of the dense and sparse data structure in order to facilitate the comparison between them.

\begin{figure}[t]
\centering
\includegraphics[width=0.9\columnwidth]{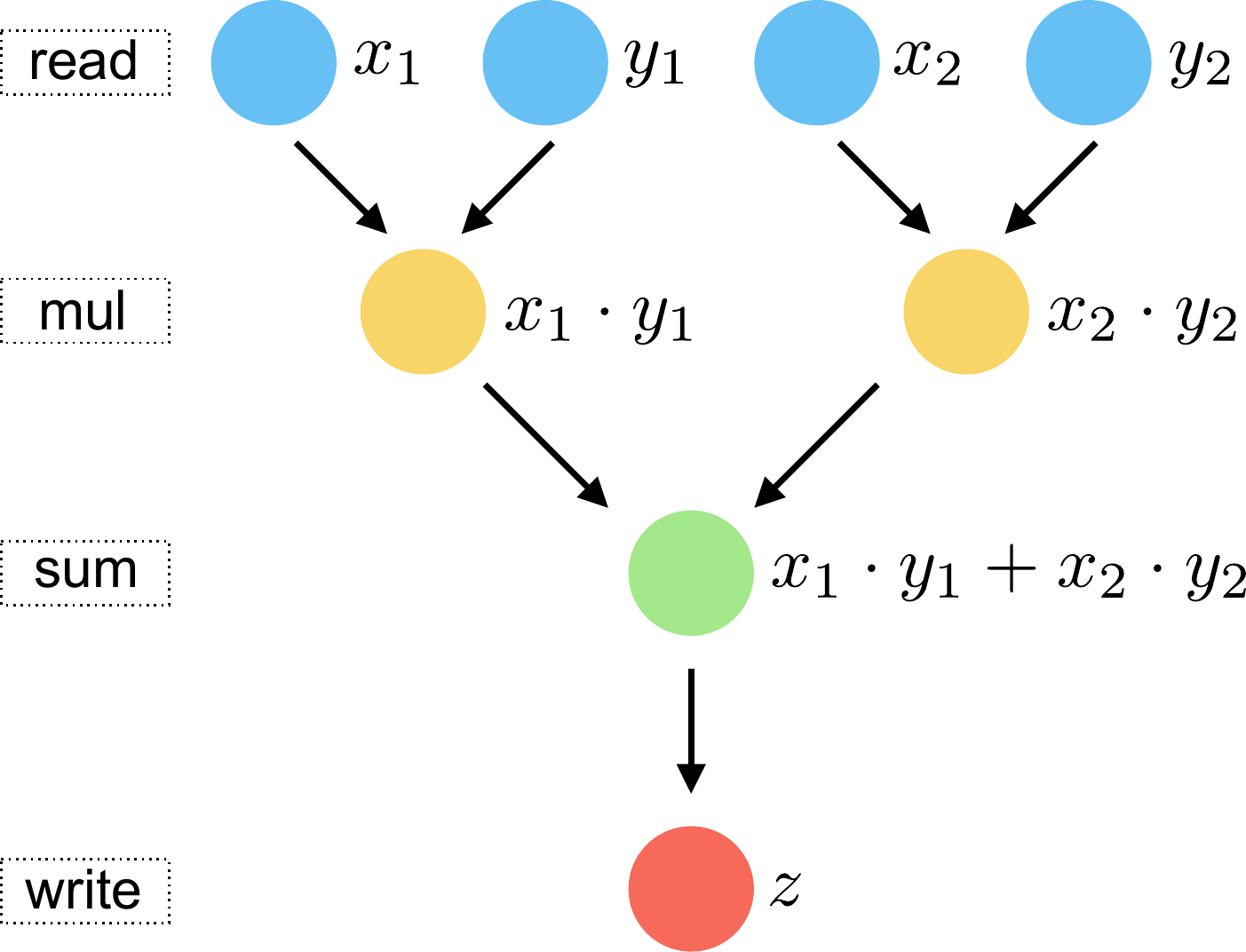}
\caption{Computational graph of a scalar product algorithm $X\cdot Y = z$ for two 2-dimensional input vectors $X, Y$. Any such algorithm can be described in terms of four elementary operations (\textit{sum}, \textit{mul}, \textit{read}, \textit{write}). These elementary operations are associated with functions $\sigma, \mu, \gamma, \delta$, which take a bit-size $b$ as input and output the energy (and/or time) cost of performing that operation. Hence, assuming that all elements have same bit-size $b$, the total energy performance of the algorithm can be determined by calculating $E = 1\sigma(b) + 2\mu(b) + 4\gamma(b) + 1\delta(b)$.}
\label{Fig: dot comp graph}
\end{figure}

\subsection{Efficiency analysis of the  dense and CSR formats}
The dense data structure stores the matrix in an $N$-long array (where $N = m\times n$) using a constant bit-size $b_{\Omega}$ for each element. Therefore, it's effective per element storage requirement is
\begin{align}
S_{\text{dense}} & = b_{\Omega}
\label{Eq: storage dense}
\end{align}
bits. 
The associated standard scalar product algorithm then has the following per element energy costs
\begin{align}
E_{\text{dense}} & = \sigma(b_o) + \mu(b_o) + \gamma(b_a) + \gamma(b_{\Omega}) + \frac{1}{n}\delta(b_o)
\label{Eq: energy dense}
\end{align}
where $b_a$ denotes the bit-size of the elements of the input vector $a\in \R^n$ and $b_o$  the bit-size of the elements of the output vector. The cost \eqref{Eq: energy dense} is derived from considering 1) loading the elements of the input vector [$\gamma(b_a)$], 2) loading the elements of the matrix [$\gamma(b_{\Omega}$)], 3) multiplying them [$\mu(b_o)$], 4) summing the multiplications [$\sigma(b_o)$], and 5) writing the result [$\delta(b_o)/n$].
We can see that both the storage and the dot product efficiency have a constant cost attached to them, despite the distribution of the elements of the matrix.

In contrast, the CSR data structure requires only 
\begin{align}
S_{\text{CSR}} & = (1-p_0)(b_{\Omega} + b_I) + \frac{1}{n}b_I
\label{Eq: storage sparse}
\end{align}
effective bits per element in order to represent the matrix, where $b_I$ denotes the bits-size of the column indices. This comes from the fact that we need in total $N(1-p_0)b_{\Omega}$ bits for representing the non-zero elements of the matrix, $N(1-p_0)b_I$ bits for their respective column indices  and $mb_I$ bits for the row pointers. 
Moreover, it requires 
\begin{align}
E_{\text{CSR}} & =  (1-p_0)(\sigma(b_o)+\mu(b_o)+\gamma(b_a)+ \gamma(b_{\Omega}) + \gamma(b_I)) \nonumber \\
  & + \frac{1}{n}\gamma(b_I) + \frac{1}{n}\delta(b_o)
\label{Eq: energy sparse}
\end{align}
units of energy per matrix element in order to perform the dot product. The expression \eqref{Eq: energy sparse} was derived from 1) loading the non-zero element values [$(1-p_0)\gamma(b_{\Omega}$)], their respective indices [$(1-p_0)\gamma(b_I)+\gamma(b_I)/n$] and the respective elements of the input vector [$\gamma(b_a)$], 2) multiplying and summing those elements [$\sigma(b_o)+\mu(b_o)$] and then 3) writing the result into memory [$\delta(b_o)/n$].

Different to the dense format, the efficiency of the CSR data structure increases as $p_0\rightarrow 1$, thus, as the number of zero elements increases. Moreover, if the matrix size is large enough, the storage requirement and the cost of performing a dot product becomes effectively 0 as $p_0\rightarrow 1$. 

For the ease of the analysis, we introduce the big $\mathcal{O}$ notation for capturing terms that depend on the shape of the matrix. In addition, we denote the following set of operations
\begin{align}
c_a & = \sigma(b_a) + \gamma(b_a) + \gamma(b_I)  \label{Eq: const cost inp} \\
c_{\Omega} & = \gamma(b_I) + \gamma(b_{\Omega}) + \mu(b_o) +\sigma(b_o) - \sigma(b_a) \label{Eq: const cost weight}
\end{align}
$c_a$ can be interpreted as the total effective cost of involving an element of the input vector in the dot product operation. Analogously can $c_{\Omega}$ be interpreted with regard to the elements of the matrix.
Hence, we can rewrite the above equations \eqref{Eq: energy dense} and \eqref{Eq: energy sparse} as follows
\begin{align}
E_{\text{dense}} & = c_a + c_{\Omega} - 2\gamma(b_I) + \mathcal{O}(1/n) \\
E_{\text{CSR}} & = (1-p_0)(c_a + c_{\Omega}) + \mathcal{O}(1/n)
\end{align}

\subsection{Efficiency analysis of the CER and CSER formats}
Following a similar reasoning as above, we can state the following theorem
\\
\begin{theorem}
Let $M \in \R^{m\times n}$ be a matrix. Let further $p_0 \in (0,1)$  be the empirical probability mass distribution of the zero element, and let $b_I \in \N$ be the bit-size of the numerical representation of a column or row index in the matrix. Then, the CER representation of $M$ requires
\begin{align}
S_{\text{CER}} & = (1-p_0)b_I + \frac{\bar{k} + \tilde{k}}{n} b_I + \mathcal{O}(1/n) + \mathcal{O}(1/N)
\label{Eq: storage cer}
\end{align}
effective bits per matrix element, where $\bar{k}$ denotes the average number of shared elements that appear per row (excluding the most frequent value), $\tilde{k}$ the average number of padded indices per row and $N = m\times n$ the total number of elements of the matrix.
Moreover, the effective cost associated to the dot product with an input vector $a\in \R^{n}$ is
\begin{align}
E_{\text{CER}} & = (1-p_0)c_a + \frac{\bar{k}}{n} c_{\Omega} + \frac{\tilde{k}}{n}\gamma(b_I) + \mathcal{O}(1/n)
\label{Eq: energy cer}
\end{align}
per matrix element, where $c_a$ and $c_{\Omega}$ are as in \eqref{Eq: const cost inp} and \eqref{Eq: const cost weight}.
\label{Thrm: strg and energy of CER}
\end{theorem}  
Analogously, we can state
\\
\begin{theorem}
Let $M$, $p_0$, $b_I$, $\bar{k}$, $c_a$, $c_{\Omega}$ be as in theorem \ref{Thrm: strg and energy of CER}. Then, the CSER representation of $M$ requires
\begin{align}
S_{\text{CSER}} & = (1-p_0)b_I + \frac{2\bar{k}}{n} b_I + \mathcal{O}(1/n) + \mathcal{O}(1/N)
\label{Eq: storage cser}
\end{align}
effective bits per matrix element, and the per element cost associated to the dot product with an input vector $a\in \R^{n}$ is
\begin{align}
E_{\text{CSER}} & = (1-p_0)c_a + \frac{\bar{k}}{n} c_{\Omega} + \frac{\bar{k}}{n}\gamma(b_I) + \mathcal{O}(1/n)
\label{Eq: energy cser}
\end{align}
\label{Thrm: strg and energy of CSER}
\end{theorem}  

The proofs of theorems \ref{Thrm: strg and energy of CER} and \ref{Thrm: strg and energy of CSER} are in the appendix.
These theorems state that the efficiency of the data structures depends on the $(\bar{k}, p_0)$ (average number of distinct elements per row - sparsity) values of the empirical distribution of the elements of the matrix. That is, these data structures are increasingly efficient for distributions that have high $p_0$ and low $\bar{k}$ values. However, since the entropy measures the effective average number of distinct values that a random variable outputs\footnote{From Shannon's source coding theorem \cite{Shannon} we know that the entropy $H$ of a random variable gives the effective average number of bits that it outputs. Therefore, we may interpret $2^H$ as the effective average number of distinct elements that a particular random variable outputs.}, both values are intrinsically related to it. In fact, from Renyi's generalized entropy definition \cite{renyi1961} we know that $p_0 \geq 2^{-H}$. Moreover, the following properties are satisfied
\begin{itemize}
\item $\bar{k}\rightarrow \min \{K-1,n\}$, as $H\rightarrow \log_2K$ or $n \rightarrow \infty$, and
\item $\bar{k}\rightarrow 0$, as $H\rightarrow 0$ or $n\rightarrow 1$.
\end{itemize}
Consequently, we can state the following corollary
\\
\begin{corollary}
For a fixed set size of unique element $|\Omega| = K$ and constant index bit-size $b_I$, the storage requirements $S$ as well as the cost of the dot product operation $E$ of the CER and CSER representations satisfy
\begin{align*}
S, E 	& \leq \mathcal{O}(1-2^{-H}) + \mathcal{O}(K/n) + \mathcal{O}(1/N) \\ 
		& = \mathcal{O}(1-2^{-H}) + \mathcal{O}(1/n)
\end{align*}
where $p_0$, $b_I$, $n$ and $N$ are as in theorems \ref{Thrm: strg and energy of CER} and \ref{Thrm: strg and energy of CSER}, and $H$ denotes the entropy of the matrix element distribution.
\label{Corollary: CER/CSER efficiency limits}
\end{corollary}

Thus, the efficiency of the CER and CSER data structures increase as the column size increases, or as the entropy decreases. Interestingly, when $n\rightarrow \infty$ both representations will converge to the same values, thus, will become equivalent. In addition, there will always exist a column size $n$ where both formats are more efficient than the original dense and sparse representations (see Fig.\ \ref{Fig: gains vs n} where this trend is demonstrated experimentally). 

\subsection{Connection between CSR, CER and CSER}
\begin{figure}[t]
\centering
\includegraphics[width=0.75\columnwidth,clip,keepaspectratio]{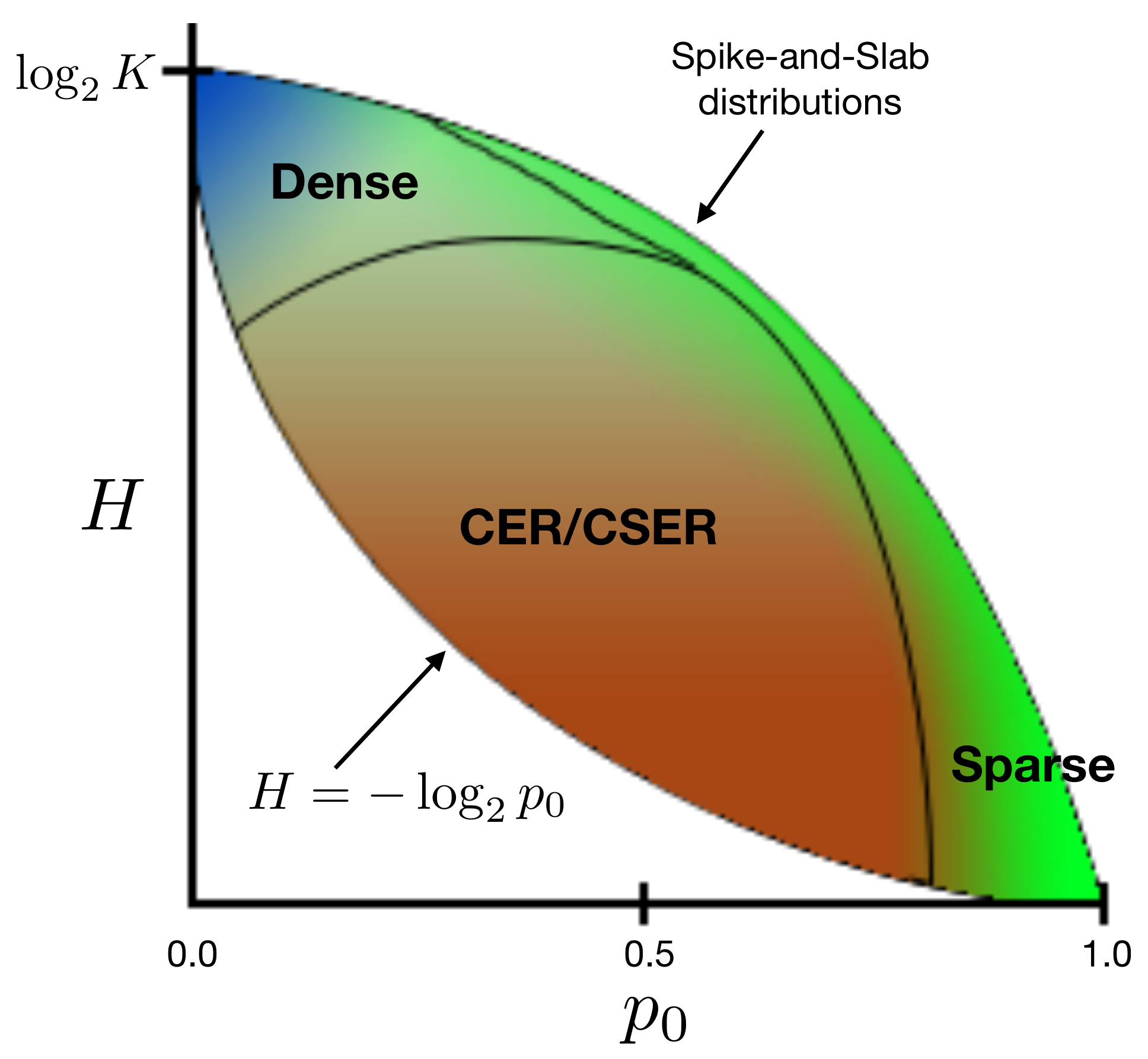}
\caption{Sketch of efficiency regions of the different data structures on the entropy-sparsity-plane ($H$ denotes the entropy and $p_0$ the sparsity). A point in the plane corresponds to a distribution of the elements of a matrix, that has respective entropy-sparsity value. The intensity of the colors reflect the degree of the efficiency of the representations. More intense red regions indicate that the CER/CSER data structures are more efficient. Respectively, the colors blue and green indicate the degree of efficiency of the dense and sparse data structures. There are two lines that constrain the set of possible distributions. The bottom line corresponds to distributions whose entropy equal their respective min-entropy (that is, where $H=-\log_2p_0$). The second line (at the most right) to the family of spike-and-slab distributions.}
\label{Fig: H vs p0 sketch}
\end{figure}

The CSR format is considered to be one of the most general sparse matrix representations, since it makes no further assumptions regarding the empirical distribution of the matrix elements. Consequently, it implicitly assumes a spike-and-slab\footnote{That is, a spike at zero with probability $p_0$ and a uniform  distribution over the non-zero elements.} distribution on them. However, spike-and-slab distributions are a particular class of low entropic (for sufficiently high sparsity levels $p_0$) distributions. In fact, spike-and-slab distributions have the highest entropy values compared to all other distributions that have same sparsity level. In contrast, as a consequence of corollary \ref{Corollary: CER/CSER efficiency limits}, the CER and CSER data structures relax this prior and can therefore efficiently represent the entire set of low entropic distributions. Hence, the CSR data structure can be interpreted as a more specialized version of the CER and CSER representations. 

This may be more evident via the following example: consider the 1st row of the matrix example from section \ref{sec: A simple but conceptual example}
\[(0 \; \; 3  \; \; 0  \; \; 2  \; \; 4  \; \; 0  \; \; 0  \; \; 2  \; \; 3  \; \; 4  \; \; 0  \; \; 4) \]
The CSER data structure would represent the above row in the following manner
\begin{align*}
\Omega: & [0,4,3,2] \nonumber \\
colI: & [ 4,  9, 11,  1,  8,  3,  7] \nonumber \\
\Omega I: & [1,2,3] \nonumber \\
\Omega Ptr: & [ 0,  3,  5,  7] \nonumber \\
rowPtr: & [ 0,  3]
\end{align*}
In comparison, the CER representation assumes that the ordering of the elements in $\Omega I$ is similar for all rows and therefore, it directly omits this array and implicitly encodes this information in the $\Omega$ array. Therefore, the CER representation can be interpreted as a more explicit/specialized version of the CSER. The representation would then be
\begin{align*}
\Omega: & [0,4,3,2] \nonumber \\
colI: & [ 4,  9, 11,  1,  8,  3,  7] \nonumber \\
\Omega Ptr: & [ 0,  3,  5,  7] \nonumber \\
rowPtr: & [ 0,  3]
\label{Eq: cer example}
\end{align*}
Similarly, the CSR representation omits the $\Omega Ptr$ array since it assumes a uniform distribution over the non-zero elements (thus, over the $\Omega$ array), and in such case all the entries in $\Omega Ptr$ would redundantly be equal to 1. Therefore, the respective representation would be
\begin{align*}
\Omega: & [3, 2, 4, 2, 3, 4, 4] \nonumber \\
colI: & [ 1,  3,  4,  7,  8,  9, 11] \nonumber \\
rowPtr: & [ 0,  7]
\end{align*} 
Consequently, the CER and CSER representations will have superior performance for all those distributions that are not similar to the spike-and-slab distributions. Figure \ref{Fig: H vs p0 sketch} displays a sketch of the regions on the entropy-sparsity plane where we expect the different data structures to be more efficient. The sketch shows that the efficiency of sparse data structures is high on the subset of distributions that are close to the right border line of the $(H,p_0)$-plane, thus, that are close to the family of spike-and-slab distribution. In contrast, dense representations are increasingly efficient for high entropic distributions, hence, in the upper-left region. The CER and CSER data structures would then cover the rest of them. Figure \ref{Fig: H vs p0} confirms this trend experimentally. 

\section{Experiments}
\label{sec: Experiments}
We applied the dense, CSR, CER and CSER representations on simulated matrices as well as on quantized neural network weight matrices, and benchmarked their efficiency with regard to the following four criteria:
\begin{enumerate}
\item
\underline{Storage requirements}:
We calculated the storage requirements according to equations \eqref{Eq: storage dense}, \eqref{Eq: storage sparse}, \eqref{Eq: storage cer} and \eqref{Eq: storage cser}. 

\item
\underline{Number of operations}:
We implemented the dot product algorithms associated to the four above data structures (pseudocodes of the CER and CSER formats can be seen in the appendix) and counted the number of elementary operations they require to perform a matrix-vector multiplication.

\item
\underline{Time complexity}:
We timed each respective elementary operation and calculated the total time from the sum of those values.

\item
\underline{Energy complexity}:
We estimated the respective energy cost by weighting each operation according to Table \ref{Tbl: Energy}. The total energy results consequently from the sum of those values. As for the case of the IO operations (read/write operations), their energy cost depend on the size of the memory the values reside on. Therefore, we calculated the total size of the array where a particular number is entailed and chose the respective maximum energy value. For instance, if a particular column index is stored using a 16 bit representation and the total size of the column index array is 30KB, then the respective read/write energy cost would be 5.0 pJ.
\end{enumerate}

In addition, we used single precision floating point representations for the matrix elements and unsigned integer representations for the index and pointer arrays. For the later, we compressed the index-element-values to their minimum required bit-sizes, where we restricted them to be either 8, 16 or 32 bits. 

Notice that we do not consider the complexity of converting the dense representation into the different formats in our experiments. This is justified in the context of neural network compression since we can apply this step a priori to the inference procedure. That is, in most real world scenarios one firstly convert the weight matrices, possibly with help of a capable computer, and then deploys the converted neural network into a resource constrained device. We are mostly interested in the resource consumption that will take place on the device. Nevertheless, as an additional side note we would like to mention that the algorithmic complexity of conversion into the CSR, CER and CSER representations is of $\mathcal{O}(N)$, that is, of the order of number of elements in the matrix.


\begin{table}[!t]
\renewcommand{\arraystretch}{1.3}
\centering
\caption{Energy values (in pJ) of different elementary operations for a 45nm CMOS process \cite{Horowitz}. We set the 8 bit floating point operations to be half the cost of a 16 bit operation, whereas we linearly interpolated the values in the case of the read and write operations.}
\begin{tabular}{|c||c|c|c|}
\hline
Op & 8 bits & 16 bits & 32 bits\\
\hline
\hline
float add & 0.2 & 0.4 & 0.9 \\
float mul & 0.6 & 1.1 & 3.7 \\
R/W ($<$8KB) & 1.25 & 2.5 & 5.0 \\
R/W ($<$32KB) & 2.5 & 5.0 & 10.0 \\
R/W ($<$1MB) & 12.5 & 25.0 & 50.0 \\
R/W ($>$1MB) & 250.0 & 5000.0 & 1000.0 \\
\hline
\end{tabular}
\label{Tbl: Energy}
\end{table}

\subsection{Experiments on simulated matrices}
\begin{figure}[t]
\centering
\includegraphics[width=\columnwidth,clip,keepaspectratio]{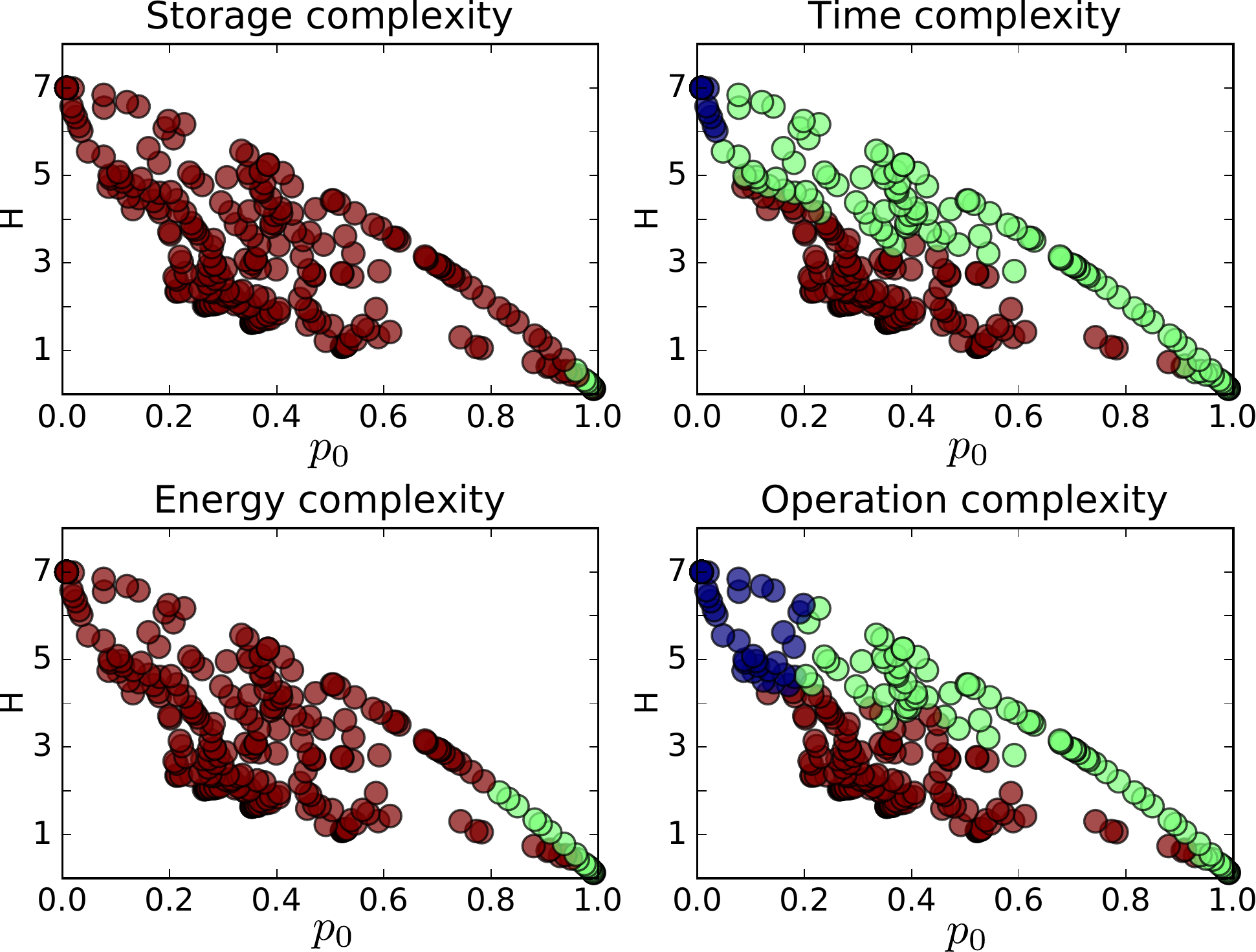}
\caption{The plots show the most efficient data structure at different points in the $H-p_0$ plane (to recall, $H$ denotes the entropy of the matrix and $p_0$ the probability of the 0 value). We compare the dense data structure (blue), the CSR format (green) and the proposed data structures (red). The colors indicate which of the three categories was the most efficient at that point in the plane. The proposed data structures tend to be more efficient in the down left region of the plane. In contrast, sparse data structures tend to be more efficiency in the upper right corner, whereas dense structures in the upper left corner. For this experiment we employed a $100\times 100$ matrix and calculated the average complexity over 10 matrix samples at each point. The size of the set of the elements was $2^7$.}
\label{Fig: H vs p0}
\end{figure}

\begin{figure}[t]
\centering
\includegraphics[width=\columnwidth,clip,keepaspectratio]{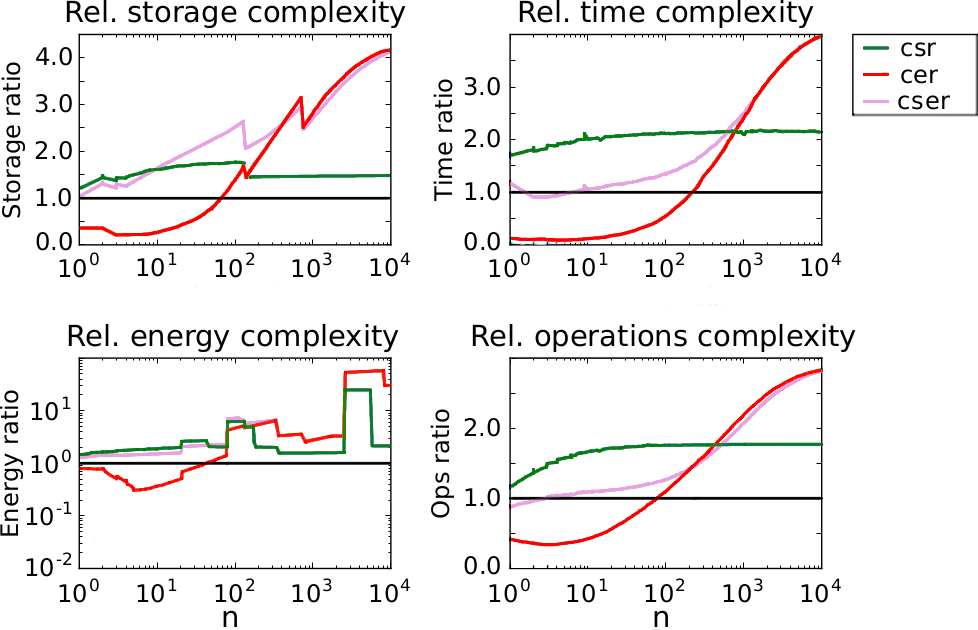}
\caption{Efficiency ratios compared to the dense data structure of the different data representations. $n$ denotes the column size. We chose a matrix with $H=4$, $p_0=0.55$ and fixed row size of 100. The results show the averaged values over 20 matrix samples. The size of the set of the elements was $2^7$. The proposed data structures tend to be more efficient as the column dimension of the matrix increases, and converge to the same value for $n\rightarrow \infty$. }
\label{Fig: gains vs n}
\end{figure}

As first experiments we aimed to confirm the theoretical trends described in Section \ref{sec: A list of crude data structures}. 

\subsubsection{Efficiency on different regions of the entropy-sparsity plane}
Firstly, we argued that each distribution has a particular entropy-sparsity value, and that the superiority of the different data structures is manifested in different regions on that plane. Concretely, we expected the dense representation to be increasingly more efficient in the upper-left corner, the CSR on the bottom-right (and along the right border) and the CER and CSER on the rest.

Figure \ref{Fig: H vs p0} shows the result of performing one such experiment. In particular, we randomly selected a point-distribution on the $(H,p_0)$-plane and sampled 10 different matrices from that distribution. Subsequently, we converted each matrix into the respective dense, CSR, CER and CSER representation, and benchmarked the performance with regard to the 4 different measures described above. We then averaged the results over these 10 different matrices. Finally, we compared the performances with each other and respectively color-coded the max result. That is, blue corresponds to points where the dense representation was the most efficient, green to the CSR and red to either the CER or CSER. As one can see, the result closely matches the expected behavior.

\subsubsection{Efficiency as a function of the column size}
As second experiment, we study the asymptotic behavior of the data structures as we increase the column size of the matrices. From corollary \ref{Corollary: CER/CSER efficiency limits} we expect that the CER and CSER data structures increase their efficiency as the number of columns in the matrix grows (thus, as $n\rightarrow \infty$), until they converge to the same point, outperforming the dense and sparse data structures. Figure \ref{Fig: gains vs n} confirms this trend experimentally with regard to all four benchmarks. Here we chose a particular point-distribution on the $(H,p_0)$-plane and fixed the number of rows. Concretely, we chose $H = 4.0$, $p_0 = 0.55$ and $m = 100$ (the later is the row dimension), and measured the average complexity of the data structures as we increased the number of columns $n\rightarrow \infty$.

As a side note, the sharp changes in the plots are due to the sharp discontinuities in the values of table \ref{Tbl: Energy}. For instance, the sharp drops in storage ratios come from the change of the index bit-sizes, e.g., from $8\rightarrow 16$ bits.  

\subsection{Compressed Neural Networks without Retraining}
\label{subsec: Compressed deep neural networks}

\begin{figure}[t]
\centering
\includegraphics[width=0.9\columnwidth,clip,keepaspectratio]{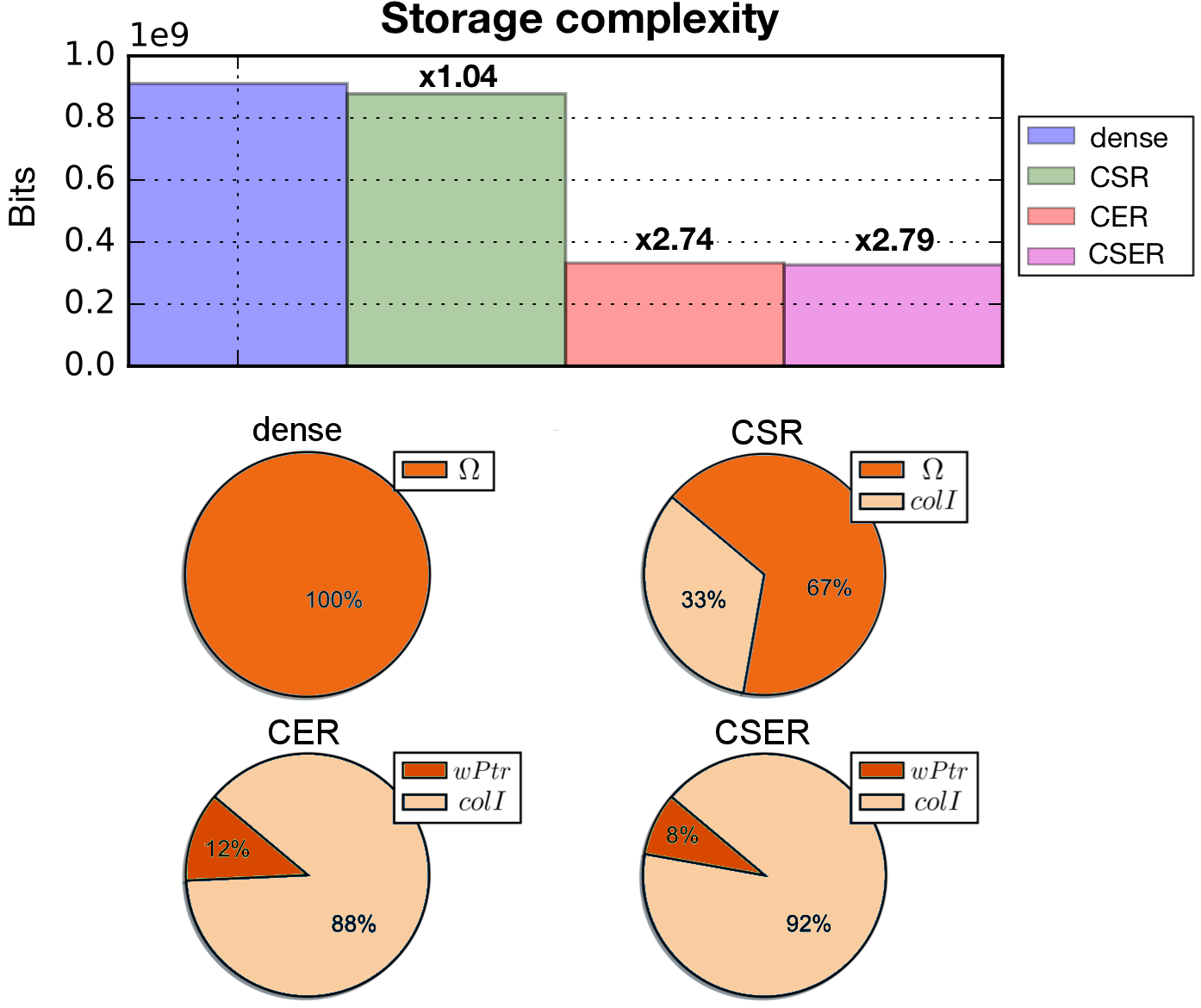}
\caption{Storage requirements of a compressed \textbf{DenseNet} \cite{DenseNet} after converting it's weight matrices into the different data structures. The weights of the network layers were compressed down to 7 bits (resulting accuracy is $77.09\%$). The plots show the over the layers averaged result. \underline{Top chart}: Compression ratio relative to the dense representation. \underline{Bottom charts}: Contribution of the different parts of the data structures to the storage requirements. For the CER/CSER formats, most of the storage goes to the column indices.}
\label{Fig: DenseNet stats strg}
\end{figure}

\begin{figure}[t]
\centering
\includegraphics[width=0.9\columnwidth,clip,keepaspectratio]{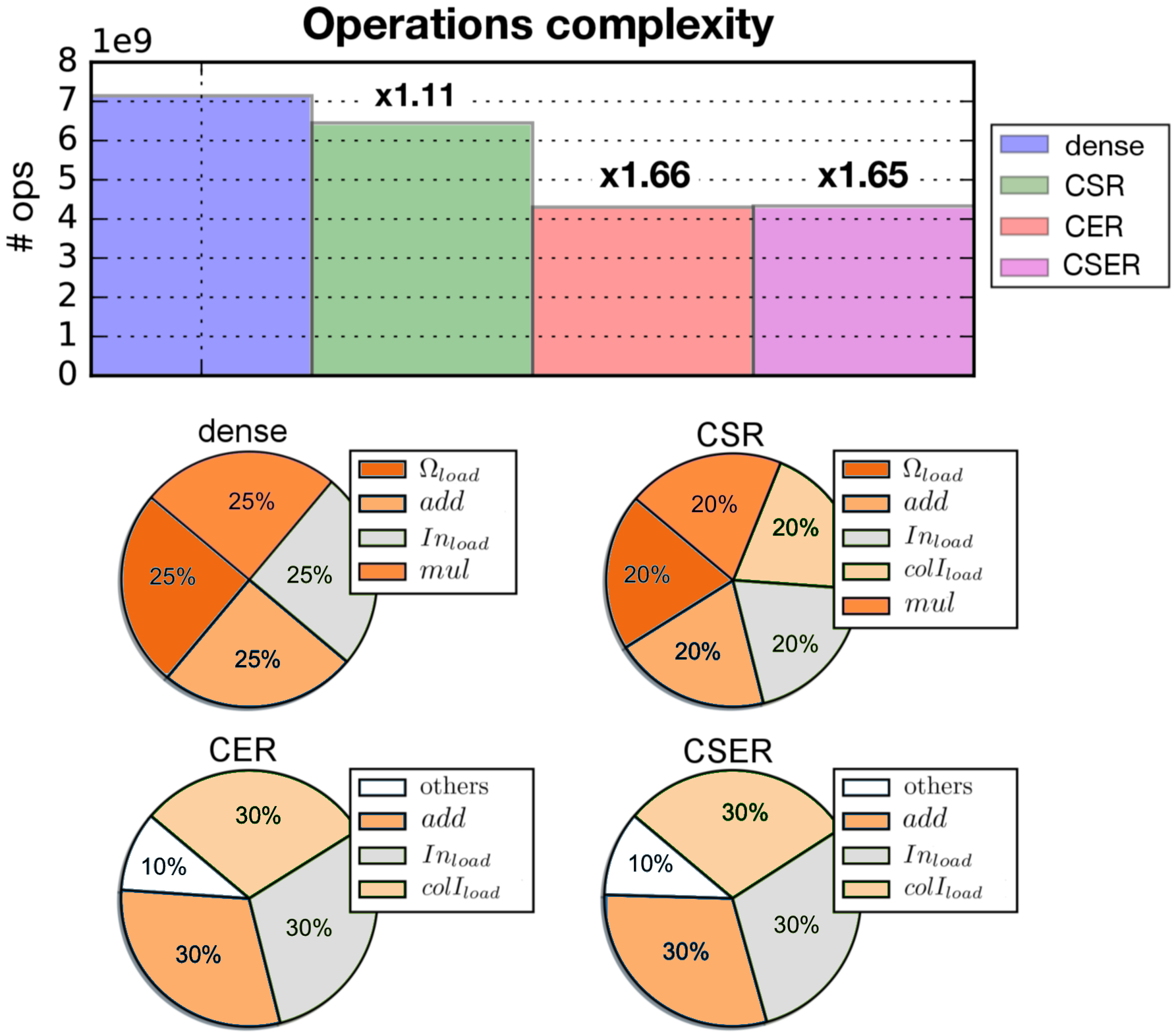}
\caption{Number of operations required to perform a dot product in the different formats for the experimental setup described in Fig.\ \ref{Fig: DenseNet stats strg} (\textbf{DenseNet}). The CER/CSER formats require less operations than the other formats, because 1) they do not need to perform as many multiplications and 2) they do not need to load as many matrix weight elements.}
\label{Fig: DenseNet stats ops}
\end{figure}

\begin{figure}[t]
\centering
\includegraphics[width=0.9\columnwidth,clip,keepaspectratio]{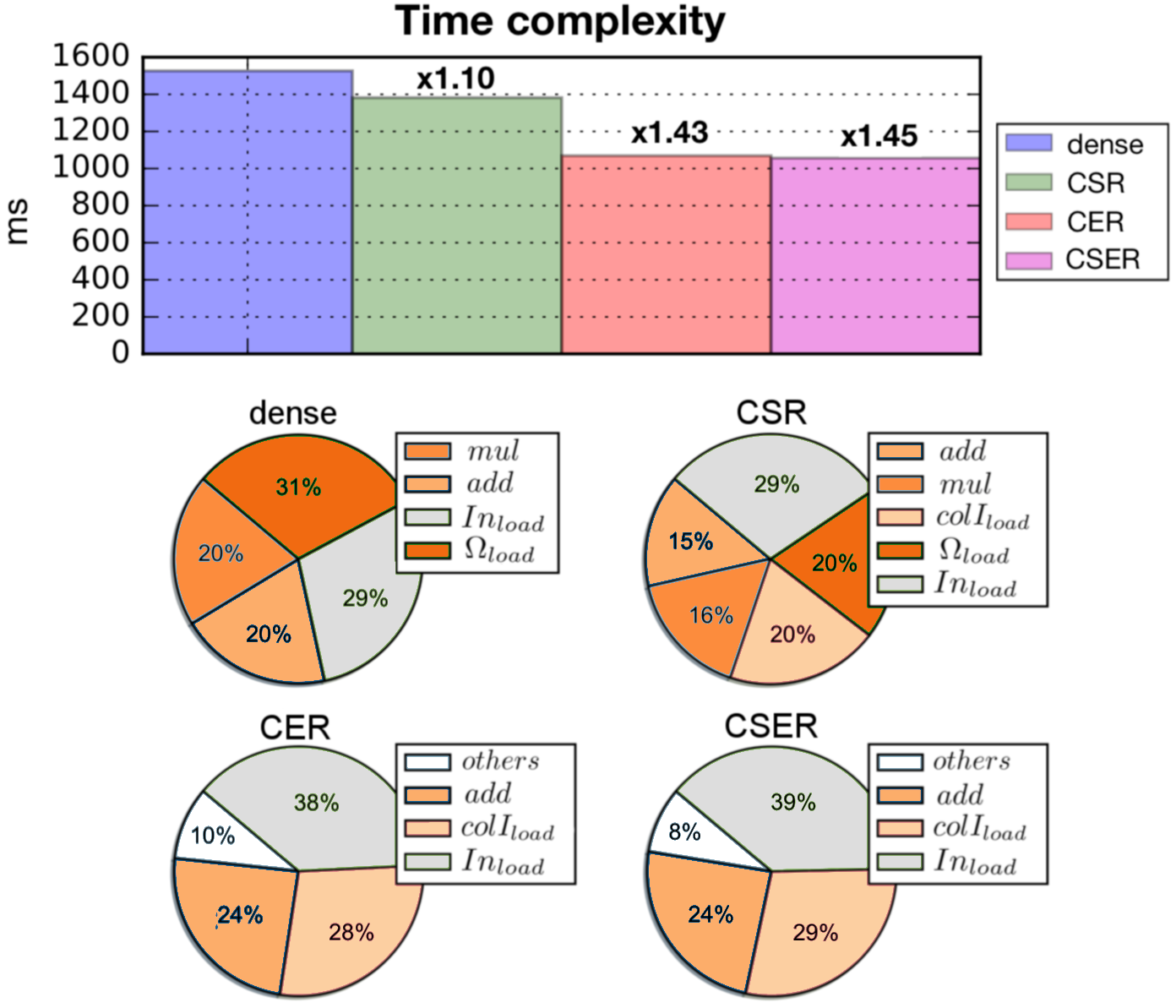}
\caption{Time cost of a dot product in the different formats for the experimental setup described in Fig.\ \ref{Fig: DenseNet stats strg} (\textbf{DenseNet}). The CER/CSER formats save time, because 1) they do not require to perform as many multiplications and 2) they do not spend as much time loading the matrix weight elements.}
\label{Fig: DenseNet stats time}
\end{figure}

\begin{figure}[t]
\centering
\includegraphics[width=0.9\columnwidth,clip,keepaspectratio]{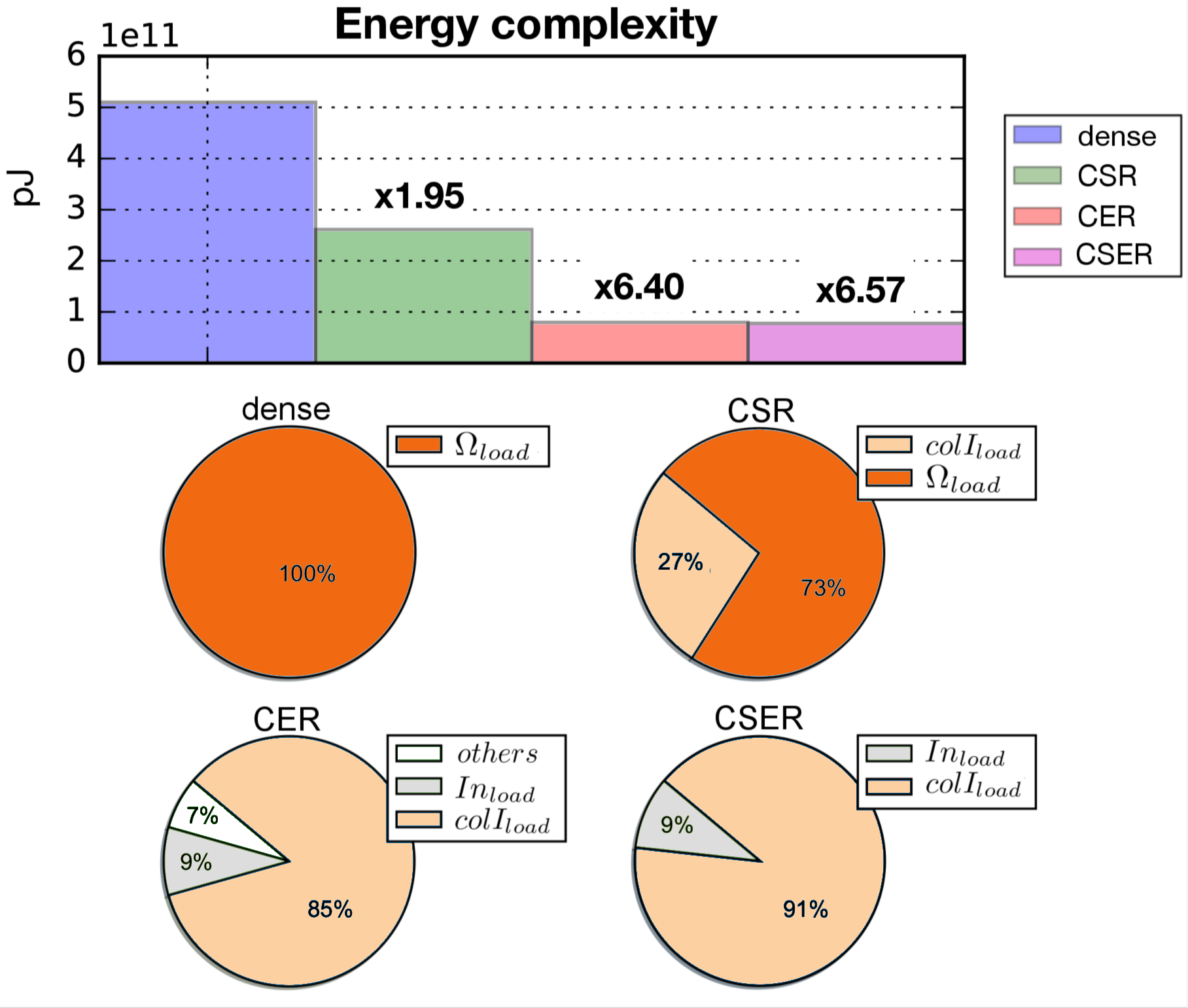}
\caption{Energy cost of a dot product in the different formats for the experimental setup described in Fig.\ \ref{Fig: DenseNet stats strg} (\textbf{DenseNet}). Performing loading operations consumes up to 3 orders more energy than sum and mul operations (see Table \ref{Tbl: Energy}). Since the CER/CSER formats need substantially less matrix weight element loading operations, they attain great energy saving compared to the dense and CSR formats.}
\label{Fig: DenseNet stats energy}
\end{figure}

As second set of experiments, we tested the efficiency of the proposed data structures on compressed deep neural networks. In particular, we benchmarked their weight matrices relative to the matrix-vector operation, after them being compressed using two different types of quantization techniques: one where retraining of the network is required (section \ref{subsec: Deep compression}) and one where it is not (section \ref{subsec: Compressed deep neural networks}). We treat them separately, since the statistics of the resulting compressed weight matrices are conditioned by the quantization applied on them.

We start by first analyzing the later case. This scenario is of particular interest since it applies to cases where one does not have access to the training data (e.g., federated learning scenario) or it is prohibited to retrain the model (e.g., limited access to computational resources). Moreover, common matrix representations, such as the dense or CSR, may fail to efficiently exploit the statistics present in these compressed weight matrices (see figure \ref{Fig: VGG class distr} and discussion in section \ref{sec:efficientNN}).

In our experiments we firstly quantized the elements of the weight matrices of the networks in a lossy manner, while ensuring that we negligible impact their prediction accuracy. Similarly to \cite{towards_limit_quantization_DNN, Universal_dnn_compression}, we applied an uniform quantizer over the range of weight values at each layer and subsequently rounded the values to their nearest quantization point. That is, for each weight matrix $W$, we calculated the range of values $[w_{\min}, w_{\max}]$ (with $w_{\min}$ being the lowest weight element value and $w_{\max}$ analogously) and inserted $K = 2^b$ equidistant points inside that range, whose values were stored in the array $\Omega$. Then, we quantized each weight element in $W$ to it's closest neighbor relative to $\Omega$ and measured the validation accuracy of the quantized network. In our experiments, we did not see any significant impact on the accuracy for all $b \geq 7$ (table \ref{Tbl: Storage Nets results}). We chose the uniform quantizer because of it's simplicity and high performance relative to other, more sophisticated quantizers such as entropy-constrained k-mean algorithms \cite{towards_limit_quantization_DNN, Universal_dnn_compression}. Finally, we lossless converted the quantized weight matrices into the different data structures and tested their efficiency with regard to the four above mentioned benchmark criteria.

\subsubsection{Storage requirements}
\begin{table}[!t]
\renewcommand{\arraystretch}{1.3}
\centering
\caption{Storage gains of different state-of-the-art neural networks after their weight matrices have been compressed down to 7 bits and, subsequently, converted into the different data structures. The gains are relative to the original dense representation of the compressed weight matrices, and they show the over the layers aggregated results. The accuracy is measured with regard to the validation set (in parenthesis we show the accuracy of the uncompressed model) of the ImageNet classification task.}
\begin{tabular}{|c||c|c||c|c|c|}
\hline
 \textbf{Storage} & Accuracy [\%] & original [MB] & CSR & CER & CSER\\
\hline
\hline
VGG16 & 68.51 (68.71) & 553.43 & x0.71 & x2.11 & x2.11 \\
\hline
ResNet152 & 78.17 (78.25) & 240.77 & x0.76 & x2.08 & x2.10 \\
\hline
DenseNet & 77.09 (77.12) & 114.72 & x1.04 & x2.74 & x2.79 \\
\hline
\end{tabular}
\label{Tbl: Storage Nets results}
\end{table}
Table \ref{Tbl: Storage Nets results} shows the gains in storage requirements of different state-of-the-art neural networks. Gains can be attained when storing the networks in CER or CSER formats. In particular, we achieve more than x2.5 savings on the DenseNet architecture, whereas in contrast the CSR data structure attains negligible gains. This is mainly attributed to the fact, that the dense and sparse representations store very inefficiently the weight element values of these networks. This is also reflected in Fig.\ \ref{Fig: DenseNet stats strg}, where one can see that most of the storage requirements for the dense and CSR representations is spent in storing the elements of the weight matrices $\Omega$. In contrast, most of the storage cost for the CER and CSER data structures comes from storing the column indices $colI$, which is much lower than the actual weight values.

\subsubsection{Number of operations}
\begin{table}[!t]
\renewcommand{\arraystretch}{1.3}
\centering
\caption{Gains attained with regard to the number of operations, time and energy cost needed for performing a matrix-vector multiplication with the compressed weight matrices of different state-of-the-art neural networks. The experiment setting and table structure is the same as in Table \ref{Tbl: Storage Nets results}. The performance gains are relative to the original dense representation of the compressed weight matrices, and they show the over the layers aggregated results.}
\begin{tabular}{|c||c||c|c|c|}
\hline
 \thead{\textbf{\#ops [G]} \\ \textbf{time [s]} \\ \textbf{energy [J]}} & original &  CSR & CER & CSER \\
\hline
\hline
VGG16 & \thead{15.08 \\ 3.37 \\ 2.70} & \thead{x0.88 \\ x0.85 \\ x0.76} & \thead{x1.40 \\ x1.27 \\ x2.37} & \thead{x1.39 \\ x1.29 \\ x2.38} \\
\hline
ResNet152 &  \thead{10.08 \\ 2.00 \\ 1.92} & \thead{x0.93 \\ x0.93 \\ x1.25 } & \thead{x1.42 \\ x1.30 \\ x3.73} & \thead{x1.41 \\ x1.31 \\ x3.74} \\
\hline
DenseNet & \thead{7.14 \\ 1.53 \\ 0.51}  & \thead{x1.11 \\ x1.10 \\ x1.95 } & \thead{x1.66 \\ x1.43 \\ x6.40 } & \thead{x1.65 \\ x1.45 \\ x6.57 }\\
\hline
\end{tabular}
\label{Tbl: Ops, time, energy Nets results}
\end{table}

Table \ref{Tbl: Ops, time, energy Nets results} shows the savings attained with regard to number of elementary operations needed to perform a matrix-vector multiplication. As one can see, we can save up to 40\% of the number of operations if we use the CER/CSER data structures on the DenseNet architecture. This is mainly due to the fact, that the dot product algorithm of the CER/CSER formats implicitly encode the distributive law of multiplications and consequently they require much less number of them. This is also reflected in Fig.\ \ref{Fig: DenseNet stats ops},  where one can see that the CER/CSER dot product algorithms are mainly performing input load ($In_{load}$), column index load ($colI_{load}$) and addition (\textit{add}) operations. Here, \textit{others} refers to any other operation involved in the dot product, such as multiplications, weight loading, writing, etc. In contrast, the dense and CSR dot product algorithms require an additional equal number of weight element load ($\Omega_{load}$) and multiplication (\textit{mul}) operations. 

\subsubsection{Time cost}
In addition, Table \ref{Tbl: Ops, time, energy Nets results} also shows that we attain speedups when performing the dot product in the new representation. Interestingly, Fig.\ \ref{Fig: DenseNet stats time} shows that most of the time is being consumed on IO's operations (that is, on \textit{load} operations). Consequently, the CER and CSER data structures attain speedups since they do not have to load as many weight elements. In addition, 20\% and 16\% of the time is spent in performing multiplications respectively in the dense and sparse representation. In contrast, this time cost is negligible for the CER and CSER representations.  

\subsubsection{Energy cost}
Similarly, we see that most of the energy consumption is due to IOs operations (Fig.\ \ref{Fig: DenseNet stats energy}).  Here the cost of loading an element may be up to 3 orders of magnitude higher than any other operations (see Table \ref{Tbl: Energy}) and therefore, we obtain up to x6 energy savings when using the CER/CSER representations (see Table \ref{Tbl: Ops, time, energy Nets results}). 

Finally, Table \ref{Tbl: NN stats} and Fig.\ \ref{Fig: H vs p0 NN layers} further justify the observed gains. Namely, Table \ref{Tbl: NN stats} shows that the effective number of shared elements per row of the network is small relative to the networks effective column dimension. To clarify, we calculated the effective number of shared elements by: 1) for all rows, calculate the number of shared weights, 2) aggregating the numbers and 3) dividing the result by the total number of rows that appear in the network. Similarly, the effective number of columns indicates the average number of columns in the network, and the effective sparsity level as well as effective entropy values indicate the over the total number of weights averaged result. Fig.\ \ref{Fig: H vs p0 NN layers} shows the distributions of the different layers of the networks on the entropy-sparsity plane where we see, that most of them lay in the regions where we expect the CER/CSER formats to be more efficient.

On a last side note we would like to comment on the alternative, compressed representations of the dense format. For instance, after quantization, we could trivially compress the weight element values down to a 7-bit representation, or apply more sophisticated entropy-coders \cite{towards_limit_quantization_DNN, Universal_dnn_compression}. Although these representation of the dense format are able to attain relatively high compression ratios, they are inefficient with regard to the dot product algorithm, since additional decoding steps are required in order to convert back the weight values into their original floating point representations. Recall, that in this case the activation values would still be represented by single precision floating point values, and quantizing them down to 7 bits would significantly harm the prediction accuracy of the network. As an example, the matrix-vector product operation of the VGG-16 architecture slowed down by about 47\% compared to the original dense representation, after we converted each weight element down into it's 7-bit representation.

\begin{figure}[t]
\centering
\includegraphics[width=0.8\columnwidth,clip,keepaspectratio]{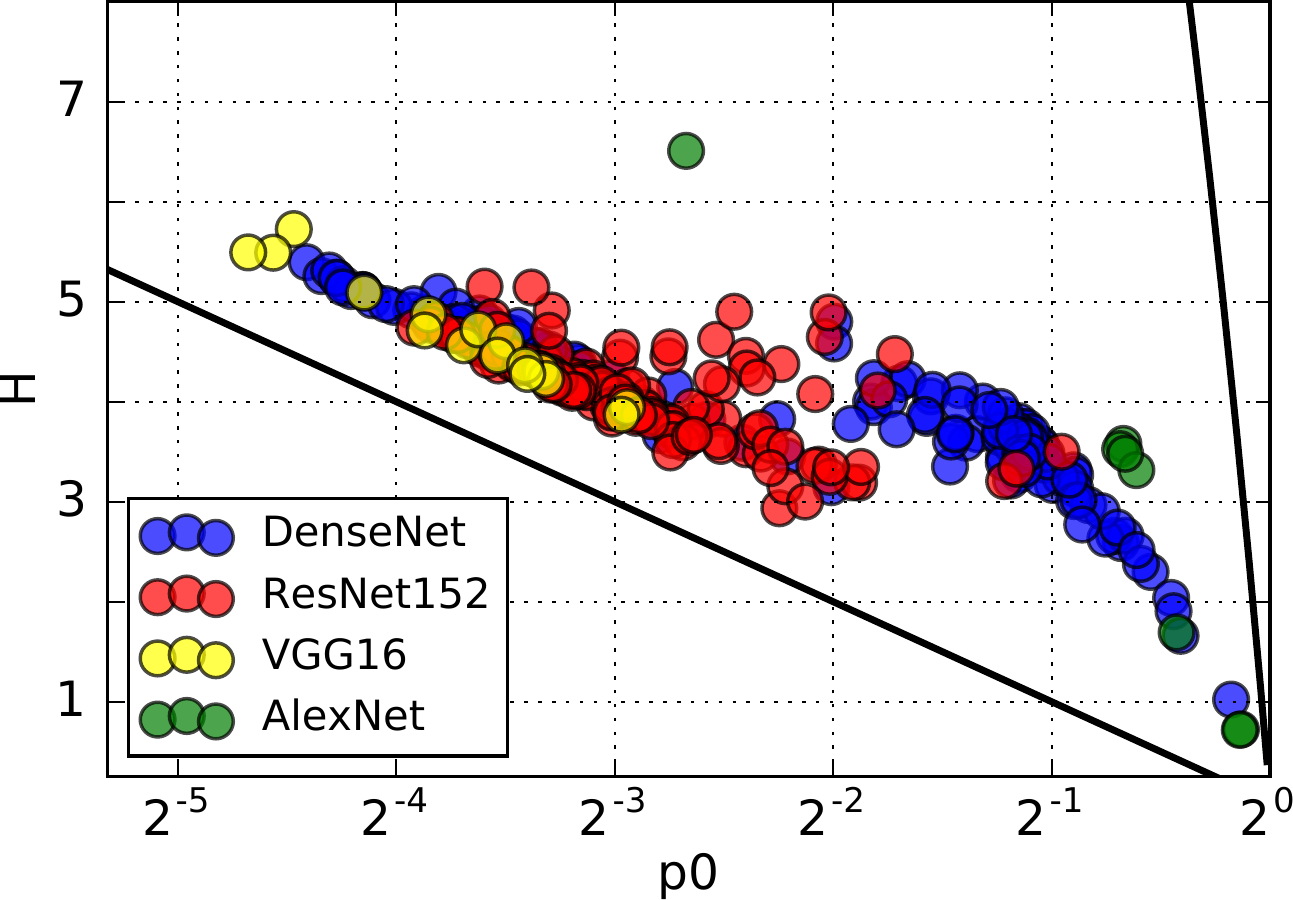}
\caption{Empirical distributions of the weight matrices of different neural network architectures after compression, displayed on the entropy-sparsity plane. As we see, most of the layers lay in the region where the CER/CSER data structures outperform the dense and sparse representation. The bottom and upper black line constrain the set of possible distributions.}
\label{Fig: H vs p0 NN layers}
\end{figure}

\begin{table}[!t]
\renewcommand{\arraystretch}{1.3}
\centering
\caption{Statistics of different neural network weight matrices taken over the entire network. $p_0$ denotes the effective sparsity level of the network, $H$ stands for the effective entropy, $\bar{k}$ represents the effective number of shared elements per row, and $n$ denotes the effective column dimension. We see that all neural networks have relatively low entropy, thus relatively low number of shared elements compared to the very high column dimensionality.}
\begin{tabular}{|c||c|c|c|c||c|}
\hline
  & $p_0$ &  $H$ & $\bar{k}$ & $n$ & $\bar{k}/n$ \\
\hline
\hline
VGG16 & 0.07 & 4.8 & 55.80 & 10311.86 & 0.01 \\
\hline
ResNet152 & 0.12 & 4.12 & 32.35 & 782.67 & 0.03 \\
\hline
DenseNet & 0.36 & 3.73 & 43.95 & 1326.93 & 0.03 \\
\hline
AlexNet \cite{deep_compression} & 0.89 & 0.89 & 18.33 & 5767.85 & 0.01 \\
\hline
\end{tabular}
\label{Tbl: NN stats}
\end{table}

\subsection{Compressed Neural Networks with Retraining}
\label{subsec: Deep compression}

\begin{figure}[t]
\centering
\includegraphics[width=\columnwidth,clip,keepaspectratio]{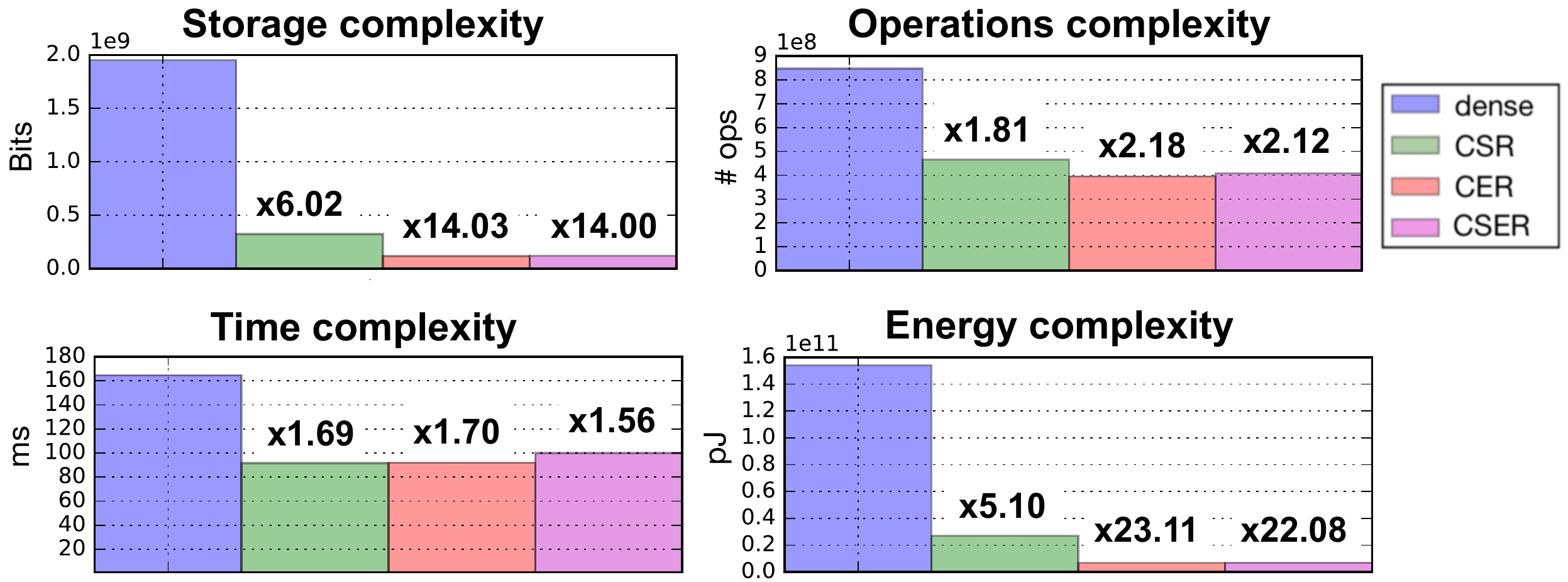}
\caption{Efficiency comparison of a compressed \textbf{AlexNet} \cite{AlexNet} after converting it's weight matrices into the different data structures and benchmarking their matrix-vector dot product operation. The network was compressed using the deep compression \cite{deep_compression} technique. The plots show the over the layers aggregated results compared to the original dense data structure.}
\label{Fig: AlexNet stats}
\end{figure}

In this section we benchmark the CER/CSER matrix representation on networks whose weight matrices have been compressed using quantization techniques where retraining was required in the process. This case is also of particular interest since highest compression gains can only be achieved if one applies such quantizations techniques on to the network \cite{deep_compression, VD_sparsifies, soft_weight_sharing, BayesianCompression, improved_BayesianCompression}.

For instance, Deep Compression \cite{deep_compression} is a technique for compressing neural networks that is able to attain high compression rates without incurring significant loss of accuracy. It is able to do so by applying a three staged pipeline: 1) prune unimportant connections by employing algorithm \cite{Learning_Weight_and_Connections}, 2) cluster the non-pruned weight values and refine the cluster centers to the loss surface and 3) employ an entropy coder for storing the final weights. Notice, that the first two stages aim to implicitly minimize the entropy of the weight matrices without incurring significant loss of accuracy, whereas the third stage lossless converts the weight matrices into low-bit representation. However, the proposed representation is based on the CSR format and, consequently, the complexity of the respective dot product algorithm remains on the same order. Concretely, the total number of operations that need to be performed is greater equal to the original CSR format. In fact, one requires specialized hardware in order to efficiently exploit this final neural network representation during inference \cite{EIE_deep_compresion}. Therefore, many authors benchmark the inference efficiency of highly compressed deep neural networks with regard to the standard CSR representation when tested on standard hardware such as CPU's and/or GPU's \cite{deep_compression, BayesianCompression, energy_aware_pruning}. However, this comes at the cost of adding redundancies since then one does not exploit step 2 of the compression pipeline.

In contrast, the CER/CSER representation become increasingly efficient as the entropy of the network is reduced, \textit{even if the sparsity level is maintained} (see figures \ref{Fig: H vs p0 sketch} and \ref{Fig: H vs p0}). Hence, it is of high interest to benchmark their efficiency on highly compressed networks and compare them to their sparse (and dense) counterparts.

As first experimental setup we chose the by the authors trained and quantized\footnote{\url{https://github.com/songhan/Deep-Compression-AlexNet}} AlexNet architecture \cite{AlexNet}, where they were able to reduce the overall entropy of the network down to 0.89 without incurring any loss of accuracy.
Figure \ref{Fig: AlexNet stats} shows the gains in efficiency when the network layers are converted into the different data structures. We see, that the proposed data structures are able to surpass the dense and sparse data structures for all four benchmark criteria. Therefore, CER/CSER data structures are much less redundant and efficient representations of highly compressed neural network models. Interestingly, the CER/CSER data structures attain up to x14 storage and x20 energy savings, which is considerably higher than the sparse counterpart. Nevertheless, we do not attain significant time gains. This is due to the fact that, in our implementations, the time cost of loading the input elements was significantly higher than any other component in the algorithm (see figure 14 in appendix). This also explains why the CSR format shows similar speedups than the CER and CSER. However, this effect can be mitigated if one applies further optimizations on the input vector, such as data reuse techniques and/or better storage management of it's values during the dot product procedure. We would also consequently expect significant gains in time performance relative to the CSR format. We will consider it in future work.

Lastly, we trained and compressed additional architectures while following a similar compression pipeline as described in \cite{deep_compression}. Concretely, we: 1) pretrained the architectures until we reached state-of-the-art accuracies, 2) sparsified the architectures using the technique proposed in \cite{VD_sparsifies}, 3) applied a uniform quantizer to the non-zero values in order to reduce their effective bit-size, finally, 4) converted the weight matrices into the different representations and benchmarked their efficiency relative to their matrix-vector product operation. In step 2) we chose \cite{VD_sparsifies} since it is the current state-of-the-art sparsification technique. In our experiments we chose to benchmark the same architectures as reported in \cite{VD_sparsifies, BayesianCompression}.  That is, an adapted version of the VGG network\footnote{http://torch.ch/blog/2015/07/30/cifar.html.} for the CIFAR-10 image classification task and the fully connected and convolutional LeNet architectures for the MNIST classification task. The respective accuracies and compression gains can be seen in tables \ref{Tbl: DC storage results} and the gains relative to the dot product complexity in table \ref{Tbl: DC ops time energy results}. As we can see, we attain significantly higher gains in all four benchmarks when we convert their weight matrices into the CER/CSER representations. In particular, we are able to attain up to x42 compression gains, x5 speedups and x90 energy gains on the VGG model.

As a last side note we want to mention again that compressing further the CSR representation by, for instance, replacing the non-zero values by their respective quantization indices (as proposed by \cite{deep_compression}), does not necessarily result in higher gains with regards to the dot product since it requires an additional decoding step per non-zero element in the process. For instance, we got only x2.89 speedups on our compressed CIFAR10-VGG model, which is less than the speedups attained by the original CSR format (x3.63). Moreover, the CER/CSER representations still attained higher gains in all other complexity measures. Concretely, we attained x33.62, x3.10 and x62.32  gains in storage, number of operations and energy respectively, which is still lower than the gains attained by the CER/CSER representations (tables \ref{Tbl: DC storage results} and \ref{Tbl: DC ops time energy results}).

\begin{table}[!t]
\renewcommand{\arraystretch}{1.3}
\centering
\caption{Storage gains of different neural networks after they have been compressed by the procedure described in section \ref{subsec: Deep compression}. The VGG model was trained on the CIFAR-10 data set and we used the same architecture as  benchmarked in \cite{VD_sparsifies, BayesianCompression}. The LeNet architectures were trained on the MNIST data set, and we took as well the same versions as benchmarked in \cite{VD_sparsifies, BayesianCompression}. The accuracy column (Acc) shows the accuracies of the compressed models, and  in parenthesis the accuracies of the pretrained models.  Finally, the sparsity column (sp) displays the ratio between the non-zero weight values and the total number of weight elements.}
\begin{tabular}{|c||c|c|c||c|c|c|}
\hline
 \textbf{Storage } & Acc [\%] & sp [\%] & orgnl [MB] & CSR & CER & CSER \\
\hline
\hline
\thead{VGG-\\CIFAR10} & \thead{90.13 \\ (91.54)}   & 4.28 & 59.91 & x17.00 & \textbf{x41.95} & x41.59 \\
\hline
\thead{LeNet-  \\ 300-100} & \thead{97.16  \\ (98.32)}  & 9.05 & 1.06 & x8.00 & \textbf{x19.52} & x18.98 \\
\hline
LeNet5 & \thead{ 98.27 \\ (99.44)}  & 1.90 & 1.722 & x35.08 & \textbf{x73.16} & x72.62 \\
\hline
\end{tabular}
\label{Tbl: DC storage results}
\end{table}

\begin{table}[!t]
\renewcommand{\arraystretch}{1.3}
\centering
\caption{Gains attained with regard to the number of operations, time and energy cost needed when benchmarking the matrix-vector multiplication of the weight matrices of the networks described in table \ref{Tbl: Storage Nets results}. The performance gains are relative to the original dense representation of the compressed weight matrices, and they display the over the layers aggregated results.}
\begin{tabular}{|c||c||c|c|c|c|}
\hline
 \thead{\textbf{\#ops [M]} \\ \textbf{time [ms]} \\ \textbf{energy [mJ]}} & original &  CSR  & CER & CSER \\
\hline
\hline
VGG-CIFAR10& \thead{878.38 \\ 208.00 \\ 139.64} & \thead{x3.71 \\ x3.63 \\ x35.41} & \thead{x5.53 \\ x5.09 \\ x89.81} & \thead{x5.43 \\ x5.10 \\ \textbf{x90.34}} \\
\hline
LeNet-300-100 &  \thead{1.065 \\ 0.25 \\ 0.02} & \thead{x9.54 \\ x9.76 \\ x14.23 } & \thead{x12.73 \\ x11.61 \\ \textbf{x54.46}} & \thead{x12.33 \\ x11.10 \\ x54.10} \\
\hline
LeNet5 & \thead{7.59 \\ 1.94 \\ 0.48}  & \thead{x3.61 \\ x3.52 \\ x60.90 } &  \thead{x4.15 \\ x3.54 \\ x87.49 } & \thead{x4.00 \\ x3.63 \\ \textbf{x96.58} }\\
\hline
\end{tabular}
\label{Tbl: DC ops time energy results}
\end{table}



\section{Conclusion}
\label{sec:conclusion}
We presented two new matrix representations, Compressed Entropy Row (CER) and Compressed Shared Elements Row (CSER), that are able to attain high compression ratios and energy savings if the distribution of the matrix elements has low entropy. We showed on an extensive set of experiments that the CER/CSER data structures are more compact and computationally efficient representations of compressed state-of-the-art neural networks than dense and sparse formats. In particular, we attained up to x42 compression ratios and x90 energy savings by representing the weight matrices of an highly compressed VGG model in their CER/CSER forms and benchmarked against the matrix-vector product operation. 

By demonstrating the advantages of entropy-optimized data formats for representing neural networks, our work opens up new directions for future research, e.g., the exploration of entropy constrained regularization and quantization techniques for compressing deep neural networks. The combination of entropy constrained regularization and quantization and entropy-optimized data formats may push the limits of neural network compression even further and also be beneficial for applications such as federated or distributed learning \cite{mcmahan2016communication, SatArXiv18}. 

Future work will also study lossy compression schemes, specially in combination with their analysis with explanation methods \cite{MonDSP18, SamITU18b}.

\ifCLASSOPTIONcaptionsoff
  \newpage
\fi

\appendices
\section{Details on neural network experiments}

\subsection{Matrix preprocessing and convolutional layers}
Before benchmarking the quantized weight matrices we applied the following preprocessing steps:

\subsubsection{Matrix decomposition}
After the quantization step it may well be that the 0 value is not included in the set of values and/or that it's not the most frequent value in the matrix. Therefore, we applied the following simple preprocessing steps: assume a particular quantized matrix $W_q\in \R^{m\times n}$, where each element $(W_q)_{ij} \in \Omega:=\{\omega_0 = 0, ..., \omega_{K-1}\}$ belong to a discrete set. Then, we decompose the matrix into the identity $W_q = (W_q - \omega_{\max}\unit) + \omega_{\max}\unit = \hat{W} + \omega_{\max}\unit$, where $\unit$ is the unit matrix whose elements are equal to 1 and $\omega_{\max}$ is the element that appears most frequently in the matrix. Consequently, $\hat{W}$ is a matrix with $0$ as it's most frequent element. Moreover, when performing the dot product with an input vector $x\in \R^n$, we only incur the additional cost of adding the constant value $c_{out} = \omega_{\max}\sum_i^n x_i$ to all the elements of the output vector. The cost of this additional operation is effectively of the order of $n$ additions and 1 multiplication for the entire dot product operation, which is negligible as long as the number of rows is sufficiently large.

\subsubsection{Convolution layers}
A convolution operation can essentially be performed by a matrix-matrix dot product operation. The weight tensor containing the filter values would be represented as a $(F_n\times(n_{ch}m_Fn_F))$-dimensional matrix, where $F_n$ is the number of filters of the layer, $n_{ch}$ the number of channels, and $(m_F,n_F)$ the height/width of the filters. Hence, the convolution matrix would perform a dot product operation with an $((n_{ch}m_Fn_F) \times n_p)$-dimensional matrix, that contains all the patches $n_p$ of the input image as column vectors. 

Hence, in our experiments, we reshaped the weight tensors of the convolutional layers into their respective matrix forms and tested their storage requirements and dot product complexity by performing a simple matrix-vector dot product, but weighted the results by the respective number of patches $n_p$ that would have been used at each layer. 

\subsection{More results from experiments}
Figures \ref{Fig:Resnet}, \ref{Fig:VGG} and \ref{Fig:AlexNet} show our results for compressed ResNet152, VGG16 and AlexNet, respectively.

\begin{figure}[t]
\centering
\includegraphics[width=0.9\columnwidth,clip,keepaspectratio]{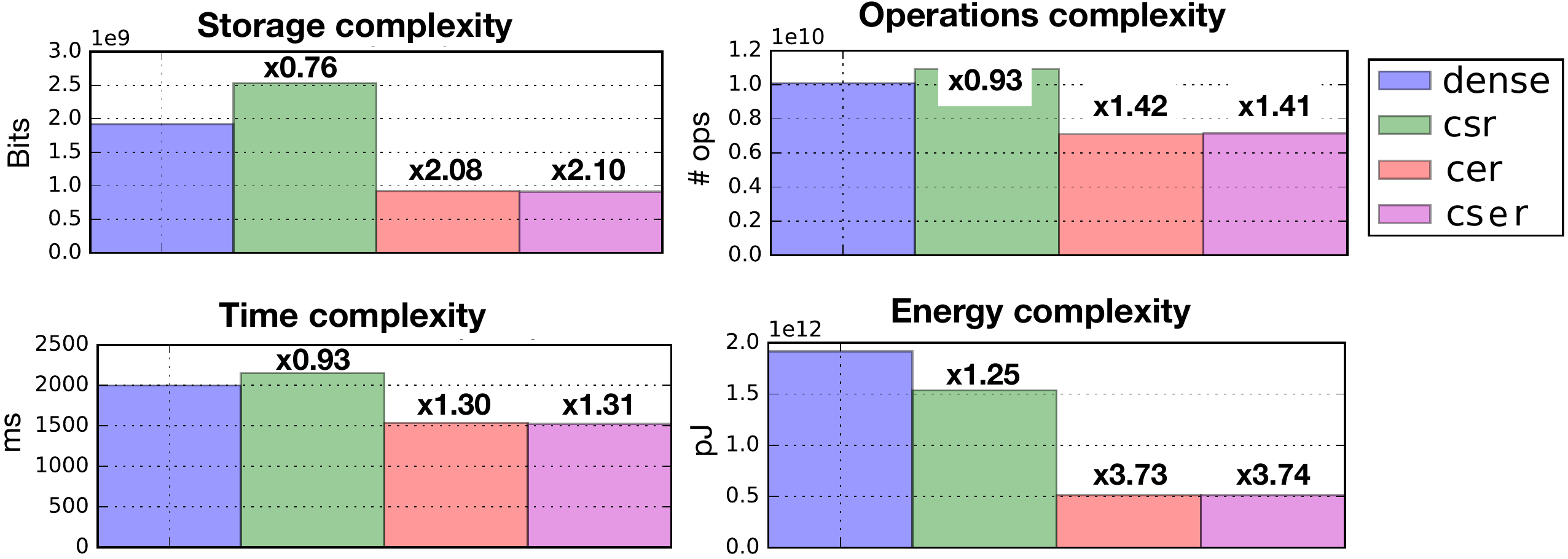}
\includegraphics[width=0.9\columnwidth,clip,keepaspectratio]{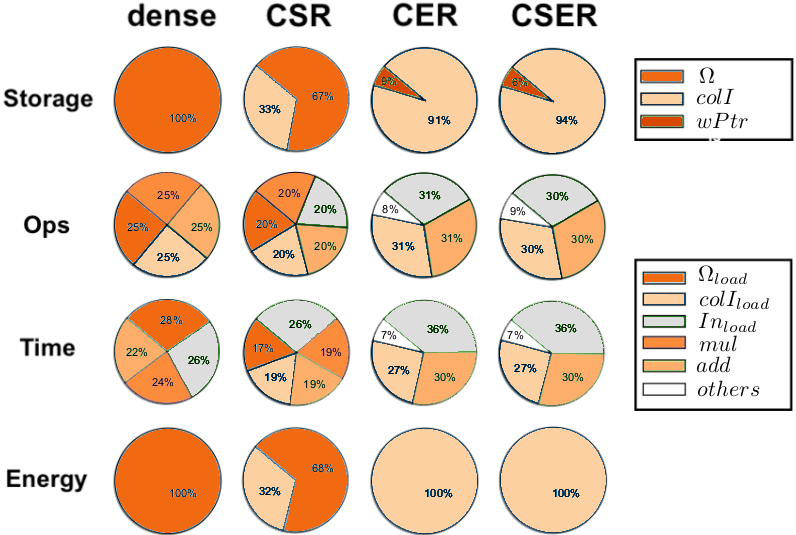}
\caption{Efficiency results from a compressed \textbf{ResNet152}. The experimental details are as described in section \ref{subsec: Compressed deep neural networks}}
\label{Fig:Resnet}
\end{figure}

\begin{figure}[t]
\centering
\includegraphics[width=0.9\columnwidth,clip,keepaspectratio]{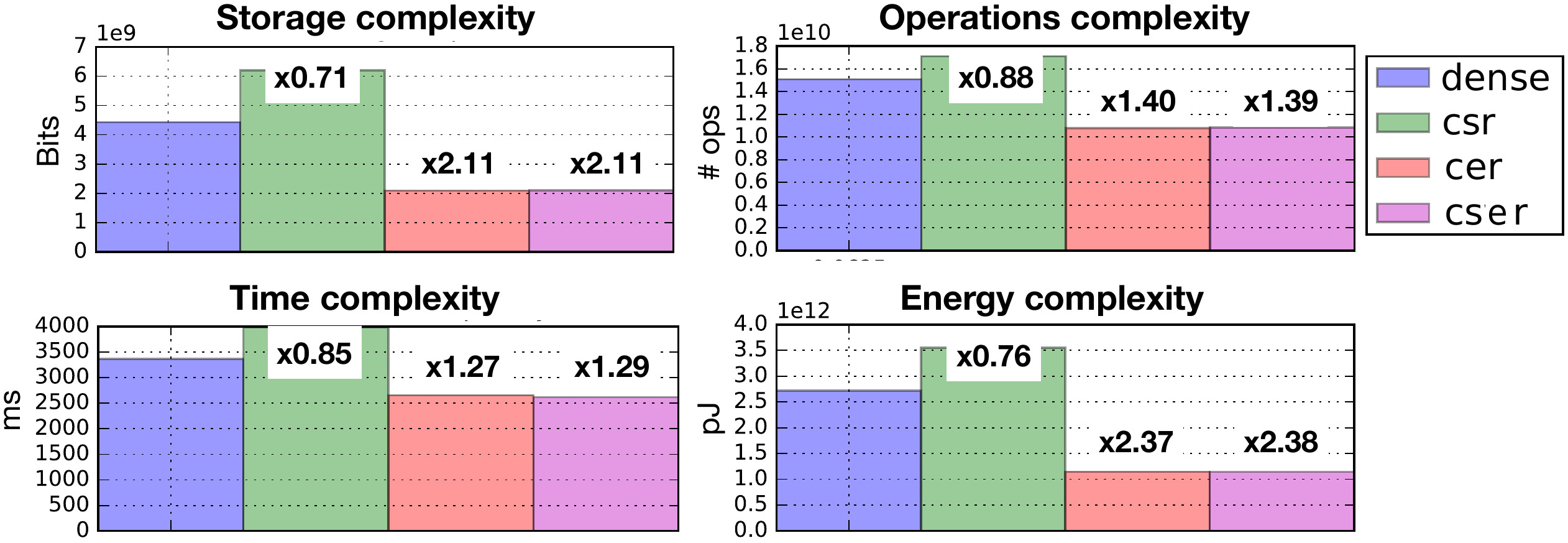}
\includegraphics[width=0.9\columnwidth,clip,keepaspectratio]{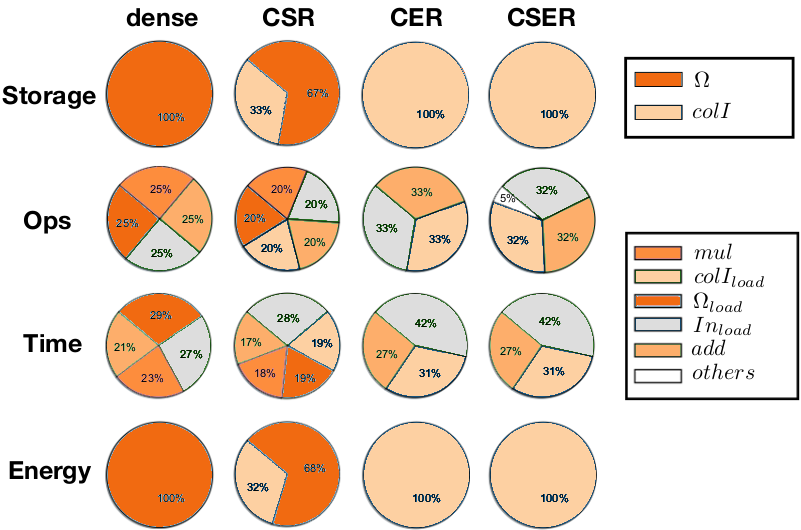}
\caption{Efficiency results from a compressed \textbf{VGG16}. The experimental details are as described in section \ref{subsec: Compressed deep neural networks}}
\label{Fig:VGG}
\end{figure}

\begin{figure}[t]
\centering
\includegraphics[width=0.9\columnwidth,clip,keepaspectratio]{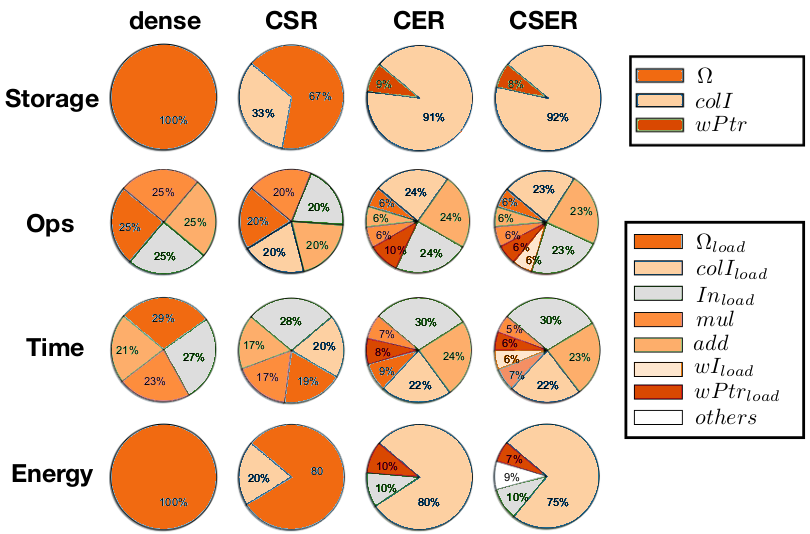}
\caption{Efficiency results of a compressed \textbf{AlexNet}. The experimental details are as described in section \ref{subsec: Deep compression}}
\label{Fig:AlexNet}
\end{figure}

\subsection{Dot product pseudocodes}
Algorithms \ref{Alg: csr dot}, \ref{Alg: cer dot} and \ref{Alg: cser dot} show the pseudocodes of the dot product algorithm of the CSR, CER and CSER data structures.

For the dense algorithm, we implemented the standard 3 loop nest algorithm \ref{Alg: dense dot}. 

We used the programming language Python\footnote{http://www.python.org} in all our experiments.

\begin{algorithm}
\caption{Dense dot product}
\label{Alg: dense dot}
\begin{algorithmic}[1]
\Procedure{dot$_{dense}$}{A,X}
\State $M, N \gets dim(A)$
\State $N, L \gets dim(X)$
\State $Y = 0 \in \R^{M\times L}$
\For{$l=0 < L$}
	\For{$m=0 < M$}
		\State $y = 0$
		\For{$n=0 < N$}
			\State $y \gets y + A[m, n]*X[n, l]$
		\EndFor
		\State $Y[m, k] \gets y$
	\EndFor
\EndFor
\State \textbf{return} $Y$
\EndProcedure
\end{algorithmic}
\end{algorithm}

\begin{algorithm}
\caption{CSR dot product}
\label{Alg: csr dot}
\begin{algorithmic}[1]
\Procedure{dot$_{csr}$}{A,X}
\State $\Omega, colI, rowPtr \gets A$ 
\State $N, L \gets dim(X)$
\State $Y = 0 \in \R^{M\times L}$
\For{$l \leq L$}
	\For{$r_{idx}=1 < len(rowPtr)$}
		\State $r_{start} \gets rowPtr[r_{idx} -1]$
		\State $r_{end} \gets rowPtr[r_{idx}]$
		\State $y = 0$
		
		\For{$i = r_{start} < r_{end}$}
			\State $I \gets colI[i]$
			\State $y \gets y + \Omega[I]*X[I,l]$
		\EndFor
		\State $Y[m, k] \gets y$
	\EndFor
\EndFor
\State \textbf{return} $Y$
\EndProcedure
\end{algorithmic}
\end{algorithm}

\begin{algorithm}
\caption{CER dot product}
\label{Alg: cer dot}
\begin{algorithmic}[1]
\Procedure{dot$_{cer}$}{A,X}
\State $\Omega, colI, wPtr, rowPtr \gets A$ 
\State $N, L \gets dim(X)$ 
\State $Y = 0 \in \R^{M\times L}$
\For{$l \leq L$}
	\State $r_{start} = 0$
	\State $w_{start} = 0$
		
	\For{$r_{idx}=1 < len(rowPtr)$}
		\State $r_{end} \gets rowPtr[r_{idx}]$
		\State $y = 0$
		\State $w_{count} = 1$
		
		\For{$w_{idx} = r_{start} + 1 < r_{end} + 1$}
			\State $w_{end} \gets wPtr[w_{idx}]$
			\State $y' = 0$
			
			\For{$i=w_{start} < w_{end}$}
				\State $I \gets colI[i]$
				\State $y' \gets y' + X[I,l]$
				\If{$w_{start}+1 = w_{end}$}
					\State $y \gets y + y' * \Omega[w_{count}]$
				\EndIf			
			\EndFor
			
			\State $w_{count} \gets w_{count} + 1$
			\State $w_{start} \gets w_{end}$
		\EndFor
		
		\State $r_{start} \gets r_{end}$
		\State $Y[m, k] \gets y$
	\EndFor
\EndFor
\State \textbf{return} $Y$
\EndProcedure
\end{algorithmic}
\end{algorithm}

\begin{algorithm}
\caption{CSER dot product}
\label{Alg: cser dot}
\begin{algorithmic}[1]
\Procedure{dot$_{cser}$}{A,X}
\State $\Omega, colI, wI, wPtr, rowPtr \gets A$
\State $N, L \gets dim(X)$ 
\State $Y = 0 \in \R^{M\times L}$ 
\For{$l \leq L$}
	\State $r_{start} = 0$
	\State $w_{start} = 0$
	\State $w_{count} = 0$
		
	\For{$r_{idx}=1 < len(rowPtr)$}
		\State $r_{end} \gets rowPtr[r_{idx}]$
		\State $y = 0$
		
		\For{$w_{idx} = r_{start} + 1 < r_{end} + 1$}
			\State $w_{end} \gets wPtr[w_{idx}]$
			\State $y' = 0$
			
			\For{$i=w_{start} < w_{end}$}
				\State $I \gets colI[i]$
				\State $y' \gets y' + X[I,l]$
			\EndFor
			
			\State $y \gets y + y' * \Omega[wI[w_{count}]]$
			\State $w_{count} \gets w_{count} + 1$
			\State $w_{start} \gets w_{end}$
		\EndFor
		
		\State $r_{start} \gets r_{end}$
		\State $Y[m, k] \gets y$
	\EndFor
\EndFor
\State \textbf{return} $Y$
\EndProcedure
\end{algorithmic}
\end{algorithm}

\section{Proof of theorems}
\subsubsection{Theorem \ref{Thrm: strg and energy of CER}}
The CER data structure represents any matrix via 4 arrays, which respectively contain: $\Omega: \; K$, $colI: \; N-\#(0)$, $wPtr: \;  \sum_{r=0}^m \bar{k}_r + \tilde{k}_r$, $rowPtr: \; m$ entries, where $K$ denotes the number of unique elements appearing in the matrix, $N$ the total number of elements, $\#(0)$ the total number of zero elements, $m$ the row dimension and finally, $\bar{k}_r$ the number of shared elements that appeared at row $r$ (excluding the 0) and $\tilde{k}_r$ the number of redundant padding entries needed to communicate at row $r$. 
\\
Hence, by multiplying each array with the respective element bit-size and dividing by the total number of elements we get
\[\frac{Kb_{\Omega}}{N} + (1-\frac{\#(0)}{N})b_I + \frac{1}{N}(\sum_{r=0}^m \bar{k}_r + \tilde{k}_r)b_I + \frac{1}{n}b_I\]
where $b_{\Omega}$ and $b_I$ are the bit-sizes of the matrix elements and the indices respectively. With $p_0 = \frac{\#(0)}{N}$ and $\bar{k} + \tilde{k} = \frac{1}{m}\sum_{r=0}^m \bar{k}_r + \tilde{k}_r$ we get equation \ref{Eq: storage cer}.
 
The cost of the respective dot product algorithm can be estimated by calculating the cost of each line of algorithm \ref{Alg: cser dot}. To recall, we denoted with $\sigma(b)$ the cost of performing a summation operation, which involved $b$ bits. $\mu(b)$ the cost of a multiplication. $\gamma(b)$ the cost of a read and $\delta(b)$ of a write operation into memory. We further denoted with $c_{oth}$ the cost of performing other types of operations. Moreover, assume an input vector (that is, $L = 1$), since the result can be trivially extended to input matrices of arbitrary size. Thus, algorithm \ref{Alg: cer dot} requires: from line 2) - 7) we assume a cost of $c_{oth}$, 8) $mc_{oth}$, 9) $m\gamma(b_I)$, 10) $mc_{oth}$, 11) $mc_{oth}$, 12) $m(\bar{k}+\tilde{k})c_{oth}$, 13) $m(\bar{k}+\tilde{k})\gamma(b_I)$, 14) $m(\bar{k}+\tilde{k})c_{oth}$, 15) $N(1-p_0)c_{oth}$, 16) $N(1-p_0)\gamma(b_I)$, 17) $N(1-p_0)(\gamma(b_a) + \sigma(b_a))$, 18) $m\bar{k}(\gamma(b_{\Omega})+ \mu(b_o) + \sigma(b_o) - \sigma(b_a))$, 19) $m(\bar{k}+\tilde{k})c_{oth}$, 20) $m(\bar{k}+\tilde{k})c_{oth}$, 21) $mc_{oth}$, 22) $m\delta(b_o)$; where $b_{\Omega}$, $b_I$ and $b_o$ are the bit-sizes of the matrix elements, the indices and output vector element respectively. Hence, adding up all above costs and replacing $c_a$ and $c_{\Omega}$ as in equations \eqref{Eq: const cost inp} and \eqref{Eq: const cost weight}, we can get the total cost of
$  c_{oth} 
 +  m(\gamma(b_I) + \delta(b_o) + 3c_{oth}) 
 +  m\bar{k}c_{\Omega}  + m\tilde{k}(\gamma(b_I)+4c_{oth}) 
  +  N(1-p_0)(c_a + c_{oth})$.
It is fair to assume that the cost $c_{oth}$ is negligible compared to the rest for highly optimized algorithms. Indeed, figures \ref{Fig: DenseNet stats time} and \ref{Fig: DenseNet stats ops} and \ref{Fig: DenseNet stats energy} show that cost of these operations contribute very little to the total cost of the algorithm. Hence, we can assume the ideal cost of the algorithm to be equal to the above expression with $c_{oth} = 0$ (which corresponds to equation \eqref{Eq: energy cer}).

\begin{flushright}
$\hfill\ensuremath{\square}$
\end{flushright}

\subsubsection{Theorem \ref{Thrm: strg and energy of CSER}}
Analogously, we can follow the same line of arguments. Namely, each array in the CSER data structure contains: $\Omega: \; K$, $colI: \; N-\#(0)$, $wI: \;  \sum_{r=0}^m \bar{k}_r$, $wPtr: \;  \sum_{r=0}^m \bar{k}_r$, $rowPtr: \; m$ entries. Consequently, by adding those terms, multiplying by their bit-size and dividing by the total number of elements $N$ we recover \eqref{Eq: storage cser}.

Each line of algorithm \ref{Alg: cser dot} induces a cost of: form line 2) - 8) we assume a cost of $c_{oth}$, 9) $mc_{oth}$, 10) $m\gamma(b_I)$, 11) $mc_{oth}$, 12) $m\bar{k}c_{oth}$, 13) $m\bar{k}\gamma(b_I)$, 14) $m\bar{k}c_{oth}$, 15) $N(1-p_0)c_{oth}$, 16) $N(1-p_0)\gamma(b_I)$, 17) $N(1-p_0)(\gamma(b_a) + \sigma(b_a))$, 18) $m\bar{k}(\gamma(b_{\Omega})+ \gamma(b_I) + \mu(b_o) + \sigma(b_o) - \sigma(b_a))$, 19) $m\bar{k}c_{oth}$, 20) $m\bar{k}c_{oth}$, 21) $mc_{oth}$, 22) $m\delta(b_o)$.
\\
Again, adding up all terms and replacing with $c_a$ and $c_{\Omega}$ then we get the total cost of
$ c_{oth} 
 +  m(\gamma(b_I) + \delta(b_o) + 3c_{oth}) 
 +  m\bar{k}(c_{\Omega} + \gamma(b_I) +4c_{oth})  
  +  N(1-p_0)(c_a + c_{oth})$ 
and with $c_{oth} = 0$ we recover equation \eqref{Eq: energy cser}.
$\ensuremath{\square}$
\bibliography{References} 
\bibliographystyle{IEEEtran}
\end{document}